\documentclass{article}

\usepackage[preprint]{neurips_2025}

\usepackage[utf8]{inputenc} 
\usepackage[T1]{fontenc}    
\usepackage{hyperref}       
\usepackage{url}            
\usepackage{booktabs}       
\usepackage{amsfonts}       
\usepackage{nicefrac}       
\usepackage{microtype}      
\usepackage{xcolor}         

\usepackage{algorithmic}
\usepackage{graphicx}
\usepackage{textcomp}
\usepackage{enumitem}

\usepackage{multirow}
\usepackage{pifont}
\usepackage{bm}
\usepackage{amsmath} 
\usepackage{amssymb}
\usepackage{bbm}
\usepackage{graphicx}

\usepackage{caption}
\usepackage{float} 
\usepackage{subcaption}
\usepackage{algorithm}
\usepackage{algorithmic}
\usepackage{wrapfig}

\usepackage[capitalize,noabbrev,nameinlink]{cleveref} 
\crefname{lem}{Lemma}{Lemmas}
\crefname{defin}{Definition}{Definitions}
\crefname{thm}{Theorem}{Theorems}

\usepackage{colortbl}
\usepackage{xcolor} 
\usepackage{soul}

\definecolor{lightgray}{gray}{0.9}

\title{Graph Unlearning via Embedding Reconstruction \\-- A Range-Null Space Decomposition Approach}

\author{%
  Hang Yin\\
  Shanghai Jiao Tong University\\
  Shanghai, China \\
  \texttt{yinhang\_SJTU@sjtu.edu.cn} \\
  \And
  Zipeng Liu\\
  Shanghai Jiao Tong University\\
  Shanghai, China \\
  \texttt{liuzipeng@sjtu.edu.cn}
  \And
  Xiaoyong Peng\\
  Shanghai Jiao Tong University\\
  Shanghai, China \\
  \texttt{pengxiaoyong0324@sjtu.edu.cn}
  \And
  Liyao Xiang\\
  Shanghai Jiao Tong University\\
  Shanghai, China \\
  \texttt{xiangliyao08@sjtu.edu.cn}
}

\begin{document}

\maketitle
\begin{abstract}
\label{Abstract}
Graph unlearning is tailored for GNNs to handle widespread and various graph structure unlearning requests, which remain largely unexplored. The GIF (graph influence function) achieves validity under partial edge unlearning, but faces challenges in dealing with more disturbing node unlearning. To avoid the overhead of retraining and realize the model utility of unlearning, we proposed a novel node unlearning method to reverse the process of aggregation in GNN by embedding reconstruction and to adopt Range-Null Space Decomposition for the nodes' interaction learning. Experimental results on multiple representative datasets demonstrate the SOTA performance of our proposed approach. 
\end{abstract}

\section{Introduction}
\label{Introduction}
Machine unlearning removes the impact of some training data from the machine learning models upon requests \cite{cao2015towards}. It is essential in many critical scenarios, such as the enforcement of laws concerning the protection of the right to be forgotten \cite{kwak2017let, pardau2018california, regulation2018general}, and the demands for model providers to revoke the negative effect of poisoned data \cite{rubinstein2009antidote, zhang2022poison}, wrongly annotated data \cite{pang2021recorrupted}, or out-of-date data \cite{wang2022causal}. The problem is particularly hard for graph data, since retraining a graph neural network (GNN) from scratch to delete a node incurs exhorbitant computational overhead. Efficient methods \cite{wu2023gif} depending on influence function have been explored, but the influence estimation of the unlearned node to the GNN is far from being exact.
\begin{figure}[htbp]
	\begin{minipage}{0.55\linewidth}
            \raggedright
		\includegraphics[height = 0.163\textheight, width = 1.15\textwidth]{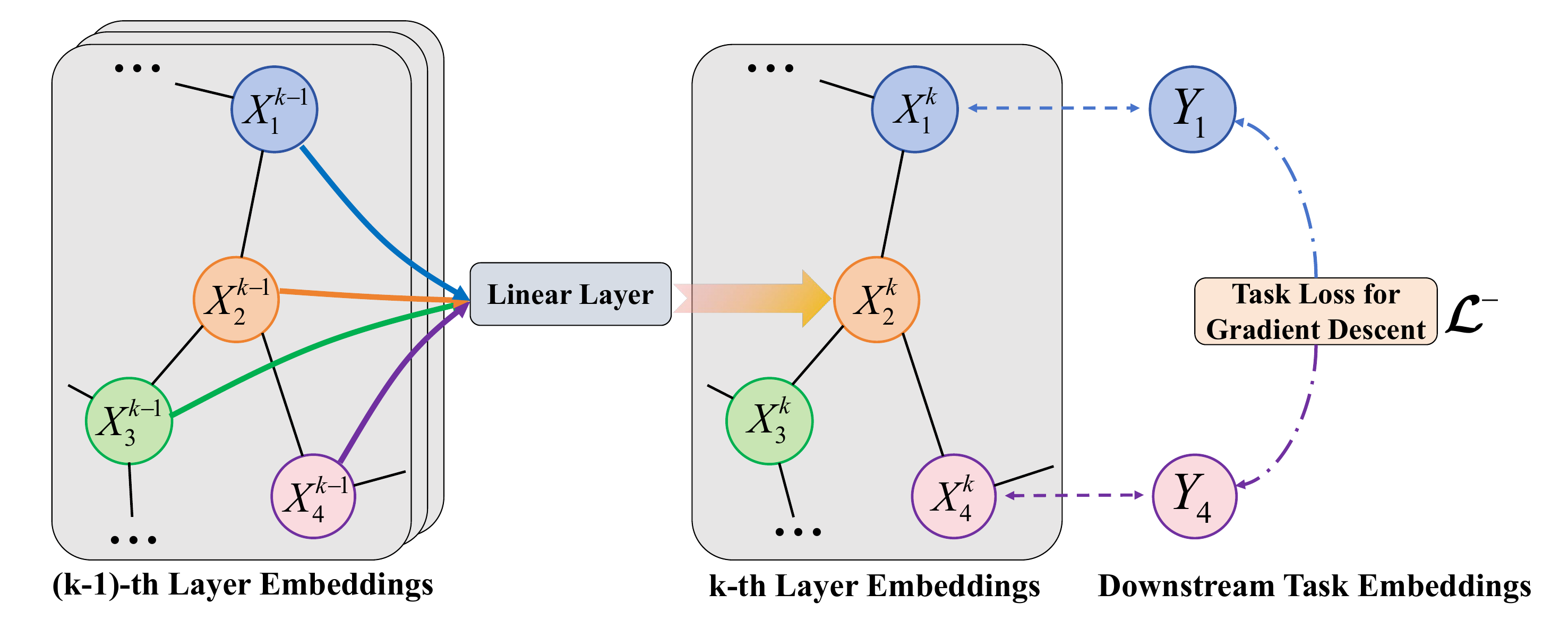}
		\caption{\centering Brief description of graph models.}
		\label{fig:Brief-GNN}
	\end{minipage}
	\begin{minipage}{0.45\linewidth}
            \raggedleft
		\includegraphics[height = 0.163\textheight, width = 0.85\textwidth]{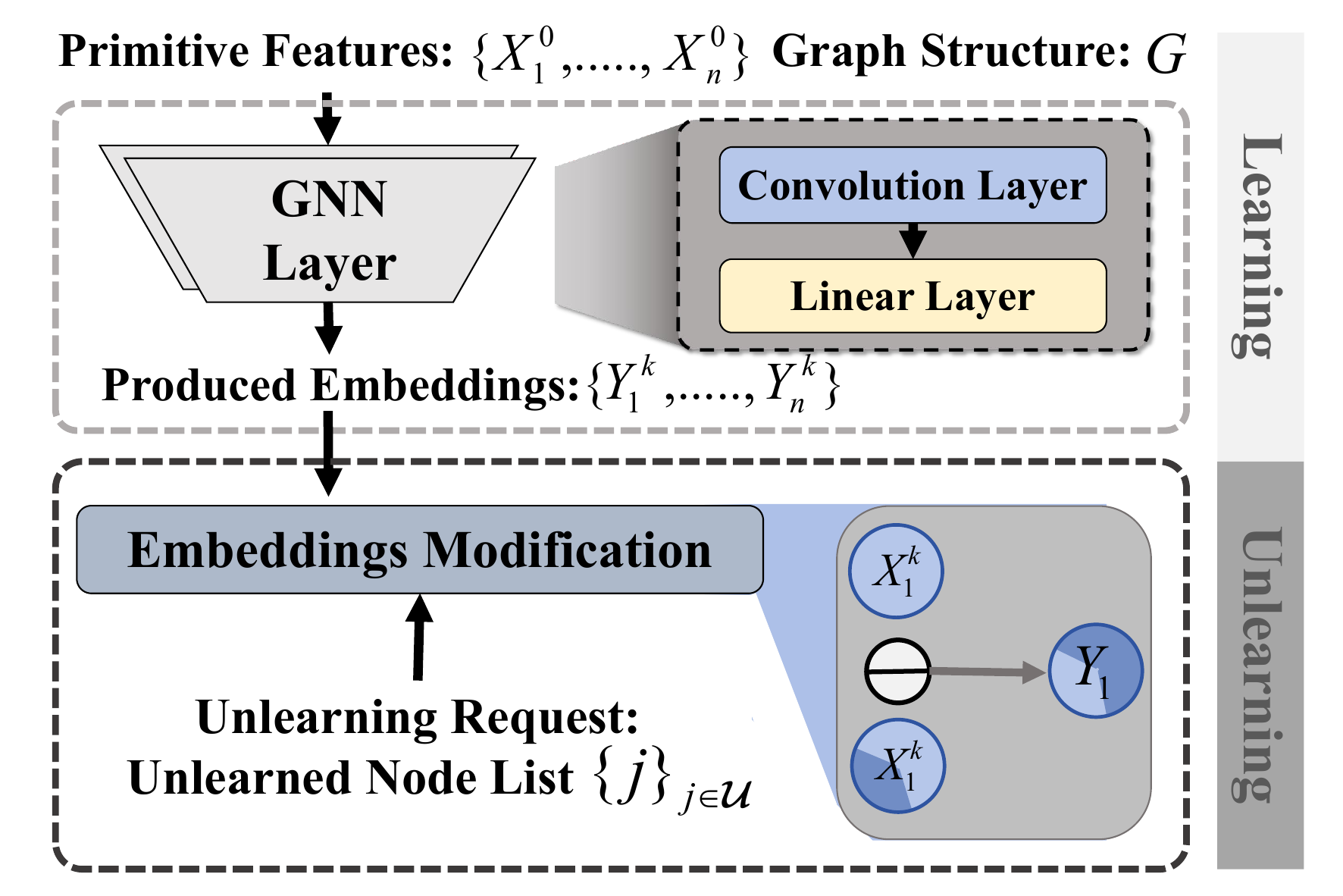}
		\caption{\centering Proposed unlearning diagram.}
		\label{fig:unlearning-diagram}
	\end{minipage}  
\end{figure}

The main challenge of graph unlearning lies in the complicated entanglement of node features woven in the GNN. Node embedding is obtained by aggregation and transformation of the representations of its neighboring nodes as shown in Fig.~\ref{fig:Brief-GNN}. There are three types of graph unlearning tasks, including \textit{Edge Unlearning}, \textit{Feature Unlearning}, and \textit{Node Unlearning}. The last type has the greatest information removal cause the node-related local neighborhood information --- including node features and adjacent edges --- is fully removed, which is the focus of our work.

Given a node as the unlearning target, one not only should remove the target node, but also offset its effect on neighbors multi-hop away. The problem is formulated as: on graph $G$, given the unlearning request $\Delta G$ indicating targeted node set $\mathcal{U}$, the goal of \textit{node unlearning} $\mathcal{M}$ is to take as inputs the original training set $\mathcal{D}_0$ and the trained GNN $f_{G}$, and to output a new GNN $\hat{f}=\mathcal{M}(\mathcal{D}_0, f_{G}, \Delta G)$ to minimize its performance discrepancy with $f_{G/\Delta G}$, the GNN trained without $\mathcal{U}$. 


The current graph unlearning works are inherently limited. SISA-based (\underline{S}harded, \underline{I}solated, \underline{S}liced, and \underline{A}ggregated) methods \cite{bourtoule2021machine, chen2022recommendation, chen2022graph} rely too much on a reasonable division of graph data into disjoint shards for sub-model retraining. The division introduces another community detection problem, letting alone the costly sub-model retraining. SISA's inference cost is also higher than others' as it requires results aggregation from all sub-models. GIF-based (Graph Influence Function) methods, represented by GIF \cite{wu2023gif}, suffers from the sensitivity problem as it only handles unlearning requests minorly changing the graph structure. Its performance heavily drops for node unlearning which causes more disturbance to the graph structure and the estimation of the influence.

To address the issue, we propose an efficient node unlearning framework based on embedding reconstruction. Instead of directly altering the GNN, we choose to modify the node embeddings to nullify the impact of the unlearned nodes, meanwhile reconstructing new embeddings to mimic those trained from scratch without $\mathcal{U}$. The key observation is that, contrary to the neighborhood aggregation of message passing in GNN learning, GNN unlearning should deduct the influence of the unlearned nodes from its neighbors. We model such an influence as a general node-wise interaction and the interaction can be reversed to reconstruct the embedding of the unlearned nodes.

However, as the GNN reduces feature dimension as it proceeds to lower layers, the reconstruction of node embedding from a lower layer to an upper layer is challenging. We resolve the issue by range-null space decomposition to let the interaction satisfy linear inverse constraints. Apart from the node-wise interaction, we guide the unlearning by a local search loss in the assumption that the embedding distributions before and after unlearning should be close. An unrolling of the gradient descent is also performed if the original GNN is trained on a downstream task.

Combining the losses, we train a rectification module for the original GNN instead of re-training a new GNN as shown in Fig.~\ref{fig:unlearning-diagram}. Thus our unlearning method is efficient with an average running time $1/40$ to $1/88$ of that of retraining, while achieving comparable model utility. We also verifies the unlearning efficacy through accuracy improvement of unlearning poisoned nodes, as well as the resistance to membership inference attacks.

Highlights of our contribution are as follows.
\begin{enumerate}[left=0pt,label=\arabic*.]
\item To the best of our knowledge, the proposed method is the first approach to solve \textit{Node Unlearning} by embedding modifications rather than the model parameters estimation. The new framework is more interpretable and helps to understand the mechanism of GNN models themselves.
\item We propose a more general and efficient algorithm tailored for GNN models. Without the assumption of gradient convergence, it is more scalable with reasonable computational overhead.
\item We conduct extensive experiments on three real-world graph datasets and four state-of-the-art GNN models to illustrate the unlearning efficiency and model utility compared with the baselines and achieve SOTA. It is also robust to adversarial data and membership inference attacks.
\end{enumerate}

\section{Related Work}
\subsection{Machine Unlearning} 

Machine unlearning aims to eliminate the influence of a subset of the training data from the trained model out of privacy protection and model security, which could also remove the influence of noisy data on model performance. Ever since Cao \& Yang \cite{cao2015towards} first introduced the concept, several methods have been proposed to address the unlearning tasks, which can be classified into two branches: exact unlearning \cite{ginart2019making, karasuyama2010multiple, bourtoule2021machine} and approximate unlearning \cite{koh2017understanding, guo2020certified, izzo2021approximate}. 

The former is aimed at creating models that perform identically to the model trained without the deleted data, or in other words, retraining from scratch, which is the most straightforward way but is computationally demanding. The \textit{SISA} (\underline{S}harded, \underline{I}solated, \underline{S}liced, and \underline{A}ggregated) approach \cite{bourtoule2021machine} partitions the data and separately trains a set of constituent models, which are afterward aggregated to form a whole model. During the procedure of unlearning, only the affected submodel is retrained with smaller fragments of data, thus greatly enhancing the unlearning efficiency. 

The latter is designed for more efficient unlearning without retraining through fine-tuning the existing model parameters. Adapting the influence function \cite{koh2017understanding} in the unlearning tasks, Guo et al. \cite{guo2020certified} proposed to unlearn by removing the influence of the deleted data on the model parameters. Unrolling SGD \cite{thudi2022unrolling} proposes a regularizer to reduce the ‘verification error’, which is an approximation to the distance between the unlearned model and a retrained-from-scratch model. Langevin Unlearning \cite{chienlangevin} leverages the Langevin dynamic analysis for the machine unlearning problem.

\subsection{Graph Unlearning}
Graph unlearning is tailored for GNNs trained with graph data. It could be divided into (1) The Shards-based method: GraphEraser \cite{chen2022graph}, GUIDE \cite{wang2023inductive}, and GraphRevoker \cite{zhang2024graph}; (2) The IF-based method: CGU \cite{chien2022certified}, GIF \cite{wu2023gif}, CEU \cite{wu2023certified}, IDEA \cite{dong2024idea}, and GST \cite{pan2023unlearning} extends the influence-function method to GNNs under the Lipschitz continuous condition and loss convergence condition; (3) Learning-based method: GNNDelete \cite{chenggnndelete} bounding edge prediction through a deletion operator, and MEGU \cite{li2024towards} achieved effective and general graph unlearning through a mutual evolution design; (4) Others: Projector \cite{cong2023efficiently} provides closed-form solutions with theoretical guarantees. 

\begin{figure*}[htbp]
	\centering
        \vspace{-12.5pt}
	\includegraphics[height = 0.41\textheight, width = 1.0\textwidth]{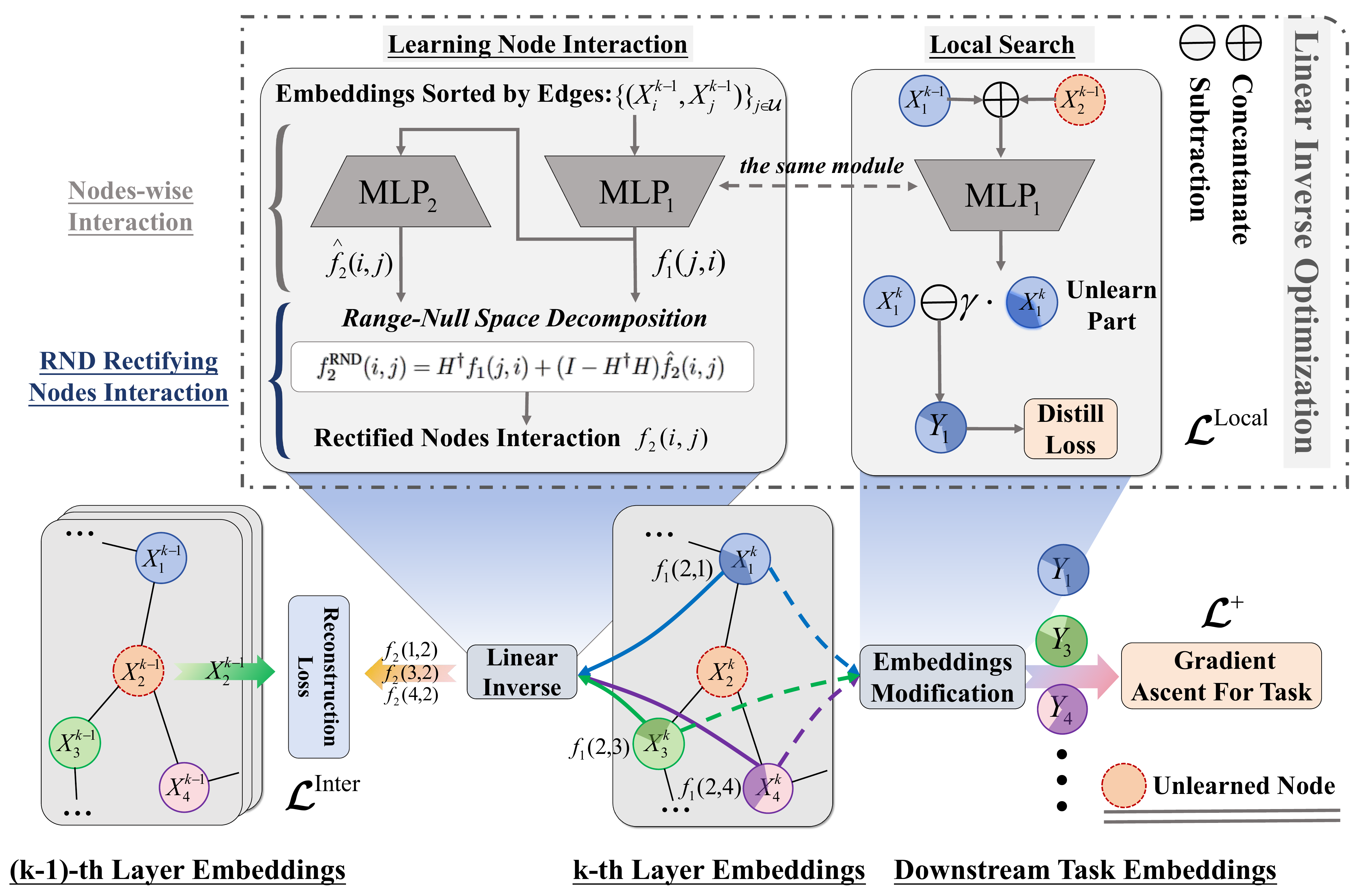}
	\caption{The illustration of our framework. Given node $2$ as an example for unlearning target, its neighbor's embeddings $X^k_i$ are modified to $Y^k_i$ as unlearned embeddings. The $f_1$ is trained to measure the interactions between nodes by reconstructing the $(k-1)$-th layer embeddings by linear inverse optimization. Range-Null Space Decomposition is employed for $f_1$'s $(k-1)$-th correspondence $f_2$ by $\mathcal{L}^\text{Inter}$. The embeddings modification is implemented by the local search with $f_1$'s interactions subtractions through $\mathcal{L}^\text{Local}$. The specific unlearning also involves the gradient ascent terms $\mathcal{L}^\text{+}$.}
	\label{fig:framework}
\end{figure*}
\vspace{-20pt}
\section{Learning Interaction via Node Embedding Reconstruction}
\label{Learning Interaction Between Nodes}
We turn the problem of graph unlearning into embedding transformation: eliminating the impact of the unlearned nodes from the retained ones so that the new embeddings are equivalent to those trained from scratch without the unlearned nodes. To achieve that, we first investigate how two nodes interact with each other on a graph.

\paragraph{Modeling node-wise interaction.}
Following the forward process of GNN, the embedding $\bm{h}^{k}_{e_i}$ of node $e_i$ in the $k$-th layer is obtained by the $(k-1)$-th layer node embeddings as: 
\begin{equation}
  \bm{h}^{k}_{e_i} = \sigma\left(\sum_{e_j\in \mathcal{N}_{e_i} \cup \{e_i\}} \alpha_{i,j} W^{k} \bm{h}^{k-1}_{e_j}\right),
  \label{eq:gnn}
\end{equation}
where $\alpha_{i,j}$ serves as the attention coefficient of nodes embeddings, $W^{k}$ is the transformation matrix of the $k$-th layer, and $\sigma$ is the activation function. Typically, $W^{k}$ projects the high-dimensional, initial representation of a node onto a lower-dimensional space. The embedding of the $k$-th layer is aggregated from those of the $(k-1)$-th layer. 

We utilize $f_1(j,i)$ to depict the influence of the unlearned node $e_j$ to the retained node $e_i$ at the $k$-th layer, given their $(k-1)$-th layer embeddings:
\begin{equation}
	\label{remove}
	f_1(j,i)=MLP_1(\bm{h}_{e_j}^{k-1},\bm{h}_{e_i}^{k-1}).
\end{equation}
The meaning of $f_1(\cdot)$ seems to be vague, but it can be interpreted as projecting the pair of embeddings of $e_i, e_j$ onto the $k$-th layer embedding space to represent $e_j$'s impact to $e_i$. We choose to use an MLP (Multi Layer Perceptrons) to represent the interaction. The goal is to remove the influence of the unlearned nodes from the $k$-th layer embedding of $e_i$:
\begin{equation}
	\tilde{\bm{h}}^{k}_{e_i}=\bm{h}^{k}_{e_i}-\sum_{j \in \mathcal{U}}f_1(j,i).
	\label{eq:unl}
\end{equation}
Thus $\tilde{\bm{h}}^{k}_{e_i}$ denotes the $k$-th layer embedding if $\mathcal{U}$ is removed.

\textbf{Remark.} According to the above Eq.~\ref{eq:gnn}, the representation of the $k$-th layer is aggregated from the representation of the $(k-1)$-th layer. Imitating its formula composition, $f_1(\cdot)$ is proposed for node interaction. MLP is employed to play the role of both linear layer $W^{k}$ and activation function $\sigma$. While the attention coefficient $\alpha_{i,j}$ could be figured out by embeddings $\bm{h}^{k-1}_{e_i}$ and $\bm{h}^{k-1}_{e_j}$. Therefore, it is enough to take MLP as a function and employ $\bm{h}^{k-1}_{e_i}$ and $\bm{h}^{k-1}_{e_j}$ as input to represent the complex interaction between nodes.

\paragraph{Embeddings reconstruction loss.} To learn the node-wise interaction, we adopt a second MLP $f_2(\cdot)$ to reconstruct the $(k-1)$-th layer embeddings from $f_1(\cdot)$. The $f_2(i,j)$ takes $f_1(j,i)$ as the input and reverses the direction of node interaction, i.e., representing how much information $e_i$ passes to $e_j$ at $(k-1)$-layer:
\begin{equation}\label{eq:mlp2}
	{f_2}(i,j)=MLP_2(f_1(j,i)).
\end{equation}
Then we aggregate the information passing from neighboring nodes to the unlearned node $e_m$ by $\sum_i{f_2(i,m)}$ and reconstruct $e_m$'s embedding at the $(k-1)$-th layer by distilling from the true $(k-1)$-th layer embedding:
\begin{equation}
	\label{L_reconst}
	\mathcal{L}^{\text{Inter}}=\frac{1}{|m|}\sum_{m \in \mathcal{U}} KL\left(\text{Norm}[\bm{h}_{e_m}^{k-1}],\text{Norm}[\sum_i{f_2(i,m)}]\right)
\end{equation}
where $\text{Norm}[\cdot]$ indicates normalization operation like $\text{Softmax}(\cdot)$, $KL(\cdot)$ represents KL divergence. The loss $\mathcal{L}^{\text{Inter}}$ is minimized over the two MLPs in reconstructing the $(k-1)$-th layer embeddings of unlearned nodes.

\paragraph{Why reconstructing the $(k-1)$-th layer?} One may question about the selection of the $(k-1)$-th layer embedding to reconstruct in modeling node-wise interaction. It is a design choice to balance the amount of information to be learned and that to be forgotten. On one hand, the $k$-th layer embedding should not be adopted in modeling the interaction since $\bm{h}^{k}_{e_j}$ has aggregated information from neighbors and contains much useful information about the graph. Removing those information would inevitably lead to GNN performance decline. On the other hand, reconstructing the representation of earlier layers is computationally inefficient as the dimension is mostly high. Considering difficulties in recovering high-dimensional features from low-dimensional ones, we choose the $(k-1)$-th layer embeddings to restore in estimating the impact between nodes. 

\section{Range-Null Space Decomposition For Rectifying Nodes Interaction }
\label{Range-Null Space Decomposition For Rectifying Nodes Interaction}
The key problem of unlearning is the lack of optimization objectives. It is unknown which embedding distributions to fit without retraining. But we are not totally ignorant. Since the unlearning set mostly occupies a minor proportion of all nodes, we expect the embedding distribution does not shift far away from the original one.

\paragraph{Establishing the linear inverse constraint between $f_1$ and $f_2$.}
Constructing $f_2$ from $f_1$ by Eq.~\ref{eq:mlp2} is intrinsically hard due to the information loss in dimension reduction. To overcome that, we employ the \textit{range-null space decomposition} for rectifying the inaccurate estimation. The technique projects the representation of a vector onto the null space (i.e., $I-H^{\dagger}H$ term) and the range space (i.e., $H^{\dagger}$ term), combining both of which could give a fine estimation of the linear inverse constrained data.

The node interaction is designed to reverse the process of GNN message passing. Since the message passing between nodes are instantiated by linear transformation and non-linear activation (Eq.~\ref{eq:gnn}), its reverse process should be represented in a similar way. That is, the linear reverse constraint between $f_1$ and $f_2$ could be established by:
\begin{equation}\label{eq:linear}
f_1(j,i)-z=H \cdot f_2(j,i),
\end{equation}
where $H$ is a linear degenerate operator, indicating the dimension reduction from the feature space of the $(k-1)$-th layer to that of the $k$-th layer. In the GNN case, $H$ indicates the weight matrix $W^k$. The term $z$ makes up for non-linear activation. To facilitate straightforward discussion of \textbf{Consistency} and \textbf{Realness} under the case: $f_1(j,i)=H \cdot f_2(j,i)$, the noisy term $z$ is eliminated below.


To estimate $f_2$ based on $f_1$, we apply range-null space decomposition as follows:
\begin{equation}
f_2^{\text{RND}}(i,j)=H^{\dagger}f_1(j,i)+(I-H^{\dagger}H){f_2}(i,j)
\label{equation7}
\end{equation}
where ${f_2}(i,j)$ is the preliminary estimation from Eq.~\ref{eq:mlp2}. Considering $H$ is dissatisfied with rank, we use $H^{\dagger}$ -- \textit{generalized inverse} of $H$ satisfying $HH^{\dagger}H\equiv H$. Through a simple linear algebraic operation, we can verify the \textbf{Consistency} of the linear inverse constraint: 
\begin{equation}
\begin{aligned}
    H\cdot f_2^{\text{RND}}(i,j)
    &=HH^{\dagger}Hf_2(i,j)+(H-HH^{\dagger}H)\hat{f_2}(i,j)\\
    &=H\cdot f_2(i,j)=f_1(j,i).
\end{aligned}
\label{equation8}
\end{equation}

Hence we replace $f_2$ in Eq.~\ref{L_reconst} with $f_2^{\text{RND}}$ in optimizing the node-wise interaction loss $\mathcal{L}^{\text{Inter}}$. For the \textbf{Realness} constraint, we scale the $f_1$ term in Eq.~\ref{eq:unl} up by a factor $\gamma=1+\frac{1}{n_g}$ where $n_g$ is the average node degree of unlearned nodes. This is because in the typical two-layer GNN, the unlearned nodes' messages spread to their $n_g$ 1-hop neighbors and approximately $n_g^2$ 2-hop neighbors on average. Considering the share of influence passing from node $e_j$ to $e_i$ as 1-hop neighbor as $1$, the share is $1/n_g$ when $e_j$ is the 2-hop neighbor of $e_i$. The final $\tilde{\bm{h}}^{k}_{e_i}$ is obtained as:
\begin{equation}
\label{unlearn_process}
\tilde{\bm{h}}^{k}_{e_i}=\bm{h}^{k}_{e_i}-\gamma\sum_{j \in \mathcal{U}}f_1(j,i)
\end{equation}

\paragraph{Local search loss.} Since the unlearning request typically involves a minor proportion of nodes, the distribution of embeddings after unlearning is assumed to be near that before unlearning. Therefore, for the retained node $p$ of high degrees, we propose to search the embedding after unlearning in a local area around their orginal embedding:
\begin{equation}
\mathcal{L}^{\text{Local}}=\sum_{p} KL(\text{Norm}[\bm{h}^{k}_{e_p}],\text{Norm}[\tilde{\bm{h}}^{k}_{e_p}]).
\end{equation}

\paragraph{Unrolling gradient descent.} If the original GNN is trained upon a downstream task, it is critical to unroll the gradient descent performed on the unlearned nodes. We take a simple gradient ascent approach to achieve that. For example, maximizing the classification loss upon unlearned nodes:
\begin{equation}
\mathcal{L}^{\text{+}}=-\sum_{j \in \mathcal{U}}CE(\text{softmax}(\tilde{\bm{h}}_{e_i}^k), y_{e_i})
\end{equation}
To sum up, the final loss at the unlearning ratio $\beta$ is:
\begin{equation}
\mathcal{L}=\beta \cdot(\mathcal{L}^{\text{+}}+\mathcal{L}^{\text{Inter}}) + (1-\beta) \cdot \mathcal{L}^{\text{Local}}.
\end{equation}
The intuition for the weight factor to be associated with the unlearning ratio is such that given a larger proportion of unlearning nodes, the learning of node-wise interaction is more important and thus a higher weight is assigned. Meanwhile, the local search term is less critical as the distribution gradually drifts away from the original one with more unlearned nodes.

Once trained, the new embedding for retained node is obtained by Eq.~\ref{unlearn_process}, which is the embedding when $\mathcal{U}$ is removed. Although we does not obtain a new GNN, but it can be equivalently obtained by the original GNN rectified by $MLP_1$. It should also be noted that our method's privacy is robust as it does not involve any retraining and thus no original node attribute participates in the unlearning.

\vspace{-10pt}
\begin{table}[h]
\centering
\caption{Statistics of the datasets.}
\setlength{\tabcolsep}{15pt}  
\resizebox{0.85\textwidth}{!}{
\begin{tabular}{c|ccccc}
\hline
Dataset  & \#Type     & \#Nodes & \#Edges & \#Features & \#Classes \\ \hline \hline
Cora     & Citation & 2,708   & 5,429   & 1,433      & 7         \\
Citeseer & Citation & 3,327   & 4,732   & 3,703      & 6         \\
CS  & Coauthor & 18,333  & 163,788 & 6,805      & 15         \\ \hline
\end{tabular}
}
\label{tab:dataset_stats}
\end{table}
\vspace{-10pt}
\section{Experiments}
\label{exp_begin}
We conduct experiments on three public graph datasets with different sizes, including Cora\footnote{https://paperswithcode.com/dataset/cora} \cite{kipf2017semi}, Citeseer\footnote{https://paperswithcode.com/dataset/citeseer} \cite{kipf2017semi}, and CS \footnote{https://pytorch-geometric.readthedocs.io/en/latest/generated/torch\_geometric.datasets.Coauthor.html} \cite{zhang2021writing}. These datasets are the benchmark datasets for evaluating the performance of GNN models for node classification task. Cora and Citeseer are citation datasets, where nodes represent the publications and edges indicate citations between two publications. CS is a coauthor dataset, where nodes are authors who are connected by an edge if they collaborate on a paper; the features represent keywords of the paper. For each dataset, we randomly split it into two subgraphs as GIF's settings \cite{wu2023gif}: a training subgraph that consists of 90\% nodes for model training and a test subgraph containing the rest 10\% nodes for evaluation. The statistics of all three datasets are summarized in Tab.~\ref{tab:dataset_stats}.

\textbf{Implementation Detail.} We use the {\tt Pytorch} framework for developing our approach. Our experiments are conducted on a workstation with an NVIDIA GeForce RTX 4090 GPUs and 24GB memory. For node unlearning tasks, we randomly delete the nodes in the training graph with an unlearning ratio $\beta$, together with the connected edges. For graph model GCN, GAT, SGC, GIN, their training basic hyperparameters are set as: learning rate $\{0.05, 0.01, 0.05, 0.01\}$ and weight decay $\{1e^{-4}, 1e^{-3}, 1e^{-4}, 1e^{-4}\}$. The $H$ matrices for GCN, GAT, and SGC are set by their convolutional linear layer weight matrices. As a special model, GIN employs two linear layers with a depth of 1 at both the beginning and end of its convolution operation. Thus, $H$ is set as the linear weight of the end one. Therefore, the $H^{\dagger}$ is obtained by {\tt numpy.linalg.pinv($H$)} through numpy packages accordingly.

\textbf{Metrics.} We evaluate the performance of our methods in terms of the following four criteria proposed in \cite{wu2023gif} and \cite{chen2022graph} comprehensively: 

\begin{enumerate}[left=0pt,label=\arabic*.]
\item \textbf{Unlearning Efficiency}: We record the running time (RT) to reflect the unlearning efficiency across different unlearning algorithms for efficiency comparison.
\item \textbf{Model Utility}: We use F1-score --- the harmonic average of precision and recall --- to measure the utility of unlearned models. This performance should be consistent with training from scratch, indicating the equivalence of unlearned models and unlearned models' utility.
\item \textbf{Unlearning Efficacy}: We propose an indirect evaluation strategy that measures model utility in the task of forgetting adversarial data.
\item \textbf{Unlearning Privacy}: We employ the membership inference attack designed for unlearning to evaluate the privacy leakage of our method by AUC values. It shows that the probability of attackers achieving the goal of attack from our method is close to randomly guessing.
\end{enumerate}

\textbf{Baselines.} We compare our method with the following approaches: \textit{first}, Retrain, the most straightforward solution that retrains the GNN model from scratch using only the remaining data. It can achieve good model utility but falls short in unlearning efficiency. \textit{Second}, GIF: an extended version of Influence Function on graph models, which considers the interaction between nodes. \textit{Third}, MEGU: the SOTA method as the learning-based baseline. More baseline comparisons are in the appendix~\ref {appendix:comp}.
\subsection{Node Unlearning Performance}\label{sec:IPUPE} 

It is noted that both the GIF method and our method are far ahead of that of retrain in unlearning efficiency as shown in Table.~\ref{tab:exp_overall}. Obviously, retraining consumes huge time and computing expenses. With the increase in graph scale, it suffers from the consumption of the $O(n^2)$ complexity of the number of nodes. Introducing more edges and nodes, the retraining cost increases significantly. This further illustrates the necessity of the problem we are exploring. 
\vspace{-10pt}

\begin{table*}[htbp]
\centering
\caption{Comparison of F1 scores and running time (RT) for different graph unlearning methods for edge unlearning with 10\% nodes deleted from the original graph. The bold indicates the best result for each GNN model on each dataset.}
\setlength{\tabcolsep}{1.8pt}  
\renewcommand\arraystretch{1}\small
\begin{tabular}{c|c|cccccc}
\hline
\multicolumn{2}{c}{Model}                             & \multicolumn{6}{c}{Dataset}  \\ \hline
\multirow{2}{*}{Backbone} & \multirow{2}{*}{Strategy} & \multicolumn{2}{c}{Cora}               & \multicolumn{2}{c}{Citeseer}           & \multicolumn{2}{c}{CS}   \\
                          &                           & F1-score                   & RT (second)        & F1-score   & RT (second)  & F1-score  & RT (second)        \\ 
                          \hline
\multirow{4}{*}{GCN}      & Retrain                   & 0.8081$\pm$0.0111          & 6.33          & 0.6973$\pm$0.0137          & 8.53          & 0.9039$\pm$0.0032          & 108.25\\
                          & GIF                       & 0.7451$\pm$0.0067          & 0.20          & 0.5876$\pm$0.0085          & 0.43          & 0.8289$\pm$0.0043          & 11.64\\ 
                          & MEGU                      & 0.7732$\pm$0.0163          & 0.19          & 0.6367$\pm$0.0128          & \textbf{0.22}          & 0.8608$\pm$0.0047          & 9.19\\
                          & Ours               & \textbf{0.8273$\pm$0.0100} & \textbf{0.16} & \textbf{0.6775$\pm$0.0090} & 0.47 & \textbf{0.9091$\pm$0.0040} &\textbf{1.23}\\ \hline
\multirow{4}{*}{GAT}      & Retrain                   & 0.8745$\pm$0.0096          & 15.72         & 0.7705$\pm$0.0049          & 35.47         & 0.9124$\pm$0.0021        & 128.37\\
                          & GIF                       & 0.8174$\pm$0.0075          & 1.39          & 0.7558$\pm$0.0046          & 1.49          & 0.8921$\pm$0.0001          & 6.74\\
                          & MEGU                      & 0.8581$\pm$0.0103          & 0.94          & 0.7905$\pm$0.0079          & 1.23          & 0.8843$\pm$0.0058          & 4.51\\
                          & Ours                       & \textbf{0.8686$\pm$0.0178} & \textbf{0.76} & \textbf{0.7813$\pm$0.0074} & \textbf{0.94} &\textbf{0.9186$\pm$0.0065} & \textbf{1.23} \\ \hline
\multirow{4}{*}{SGC}      & Retrain                   & 0.8029$\pm$0.0180          & 6.63          & 0.6949$\pm$0.0149          & 8.29          & 0.8929$\pm$0.0177          & 110.48         \\
                          & GIF                       & 0.7456$\pm$0.0071          & 0.20          & 0.5867$\pm$0.0088          & \textbf{0.22}          & 0.8334$\pm$0.0035          & 12.04          \\
                          & Ours                       & \textbf{0.8184$\pm$0.0043} & \textbf{0.14} & \textbf{0.6589$\pm$0.0056} & 0.40 & \textbf{0.9026$\pm$0.0049} & \textbf{0.96} \\ \hline
\multirow{4}{*}{GIN}      & Retrain                   & 0.8184$\pm$0.0102          & 8.48          & 0.7375$\pm$0.0165          & 25.97          & 0.8871$\pm$0.0031          & 115.34         \\
                          & GIF                       & 0.7517$\pm$0.0199          & 0.79          & 0.6849$\pm$0.0173          & 1.67          & 0.8768$\pm$0.0119          & 2.33          \\ 
                          & Ours                       & \textbf{0.8229$\pm$0.0073} & \textbf{0.30} & \textbf{0.7465$\pm$0.0056} & \textbf{0.53} & \textbf{0.8962$\pm$0.0082} & \textbf{2.02} \\ \hline
\end{tabular}
\label{tab:exp_overall}
\end{table*}
\vspace{-10pt}

Compared with retraining, our method achieves almost equivalent performance. The GIF method is more effective in solving the problem of deleting edges, as our experiments. It falls in node unlearning tasks as the neighborhood around unlearned nodes is all removed. However, GIF is good in scalability; it could be adapted to multiple backbones, and so does ours. Thus, GIF is treated as the main baseline for comparison.

The efficiency of both our method and GIF is comparable on Cora and Citeseer. With the graph scale expansion, GIF needs more iterations for convergence of a more complicated Hessian matrix, which leads to the performance gap in CS. However, ours only includes the low-consumption loss function training and the straightforward `subtraction' operations in the actual unlearning process. Its computational cost increases almost linearly with the number of nodes. The MEGU's results of GCN and GAT are also included. Its performance is better than GIF in a mass, but there is still a gap compared with our proposed method. 

\subsection{Unlearning Efficacy}\label{sec:unaware}
If the targets of unlearning are training nodes with wrong labels (i.e., the adversarial data), then intuitively, the model performance will be improved after unlearning. Accordingly, adversarial samples are constructed, and we carried out unlearning efficacy verification. We randomly shift labels of some training nodes (e.g. $y_{e_i}$), that is, their shifted labels are equal to the initial category number plus 1 (e.g. $(y_{e_i}+1)\,\text{mod}\, n_c$), while the results are modulo the number of categories (e.g. $n_c$).

\begin{figure}[ht]
    \centering
    \vspace{-10pt}
    \begin{minipage}[b]{0.245\textwidth}
        \centering
        \includegraphics[height = 0.17\textheight, width=1.14\textwidth]{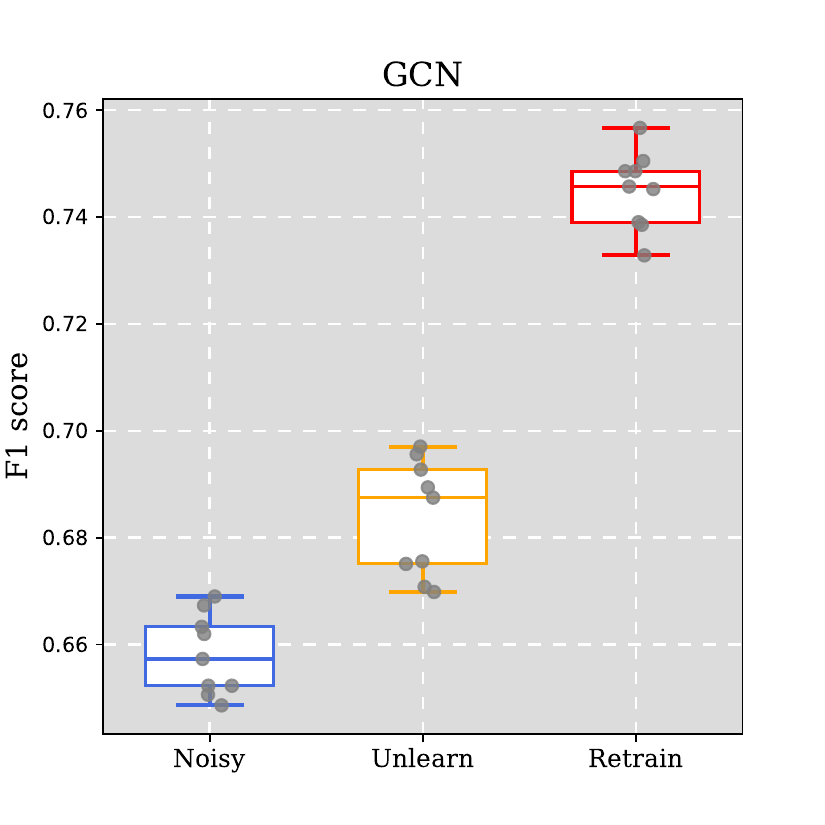}
        \put(-120,30){\rotatebox{90}{\textbf{(a) Cora}}}
    \end{minipage}
    \hfill
    \begin{minipage}[b]{0.245\textwidth}
        \centering
        \includegraphics[height = 0.17\textheight, width=1.14\textwidth]{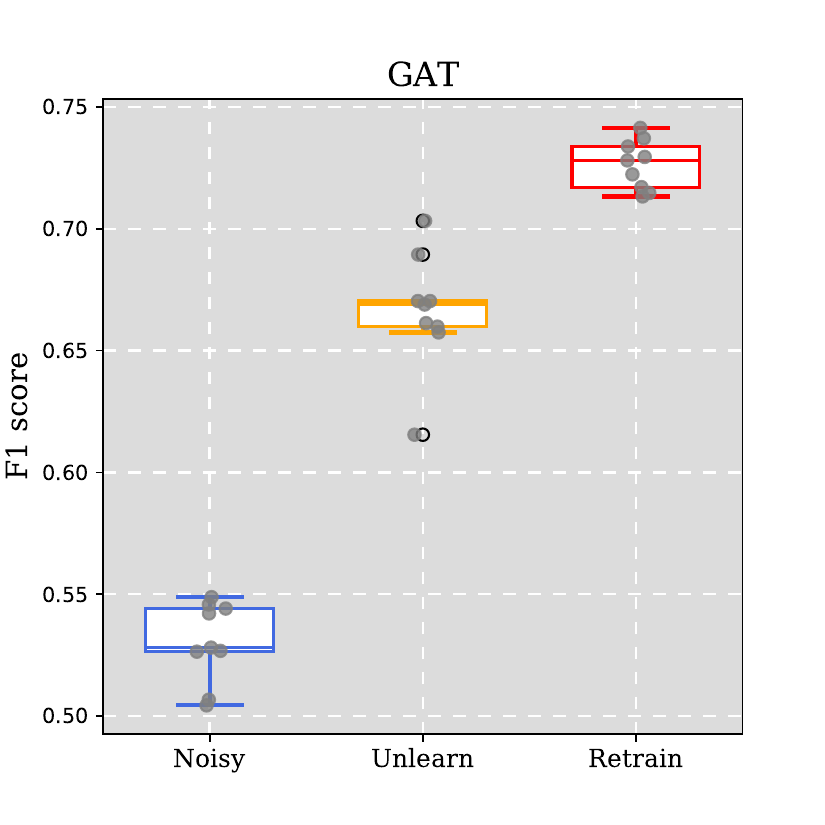}
    \end{minipage}
    \hfill
    \begin{minipage}[b]{0.245\textwidth}
        \centering
        \includegraphics[height = 0.17\textheight, width=1.14\textwidth]{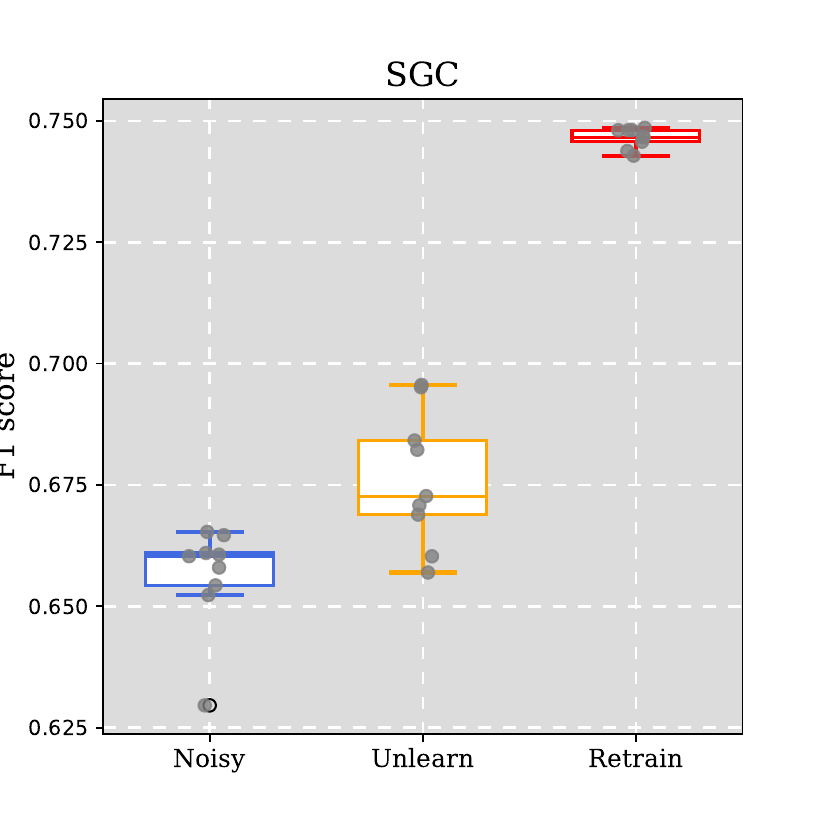}
    \end{minipage}
    \hfill
    \begin{minipage}[b]{0.245\textwidth}
        \centering
        \includegraphics[height = 0.17\textheight, width=1.14\textwidth]{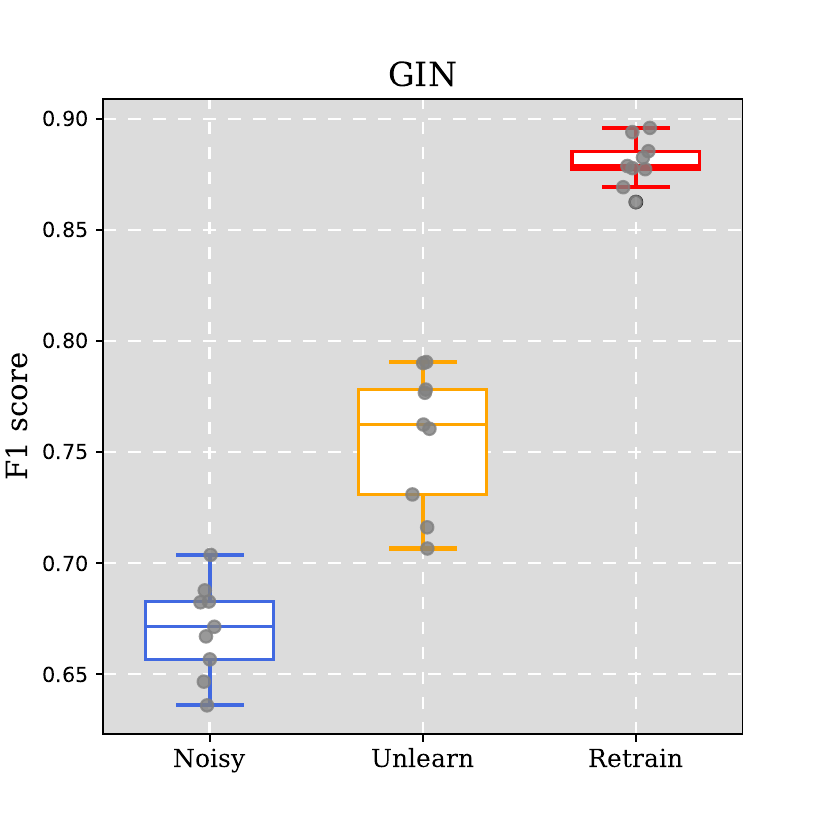}
    \end{minipage}
    \vspace{-14pt}
    
    \begin{minipage}[b]{0.245\textwidth}
        \centering
        \includegraphics[height = 0.17\textheight, width=1.14\textwidth]{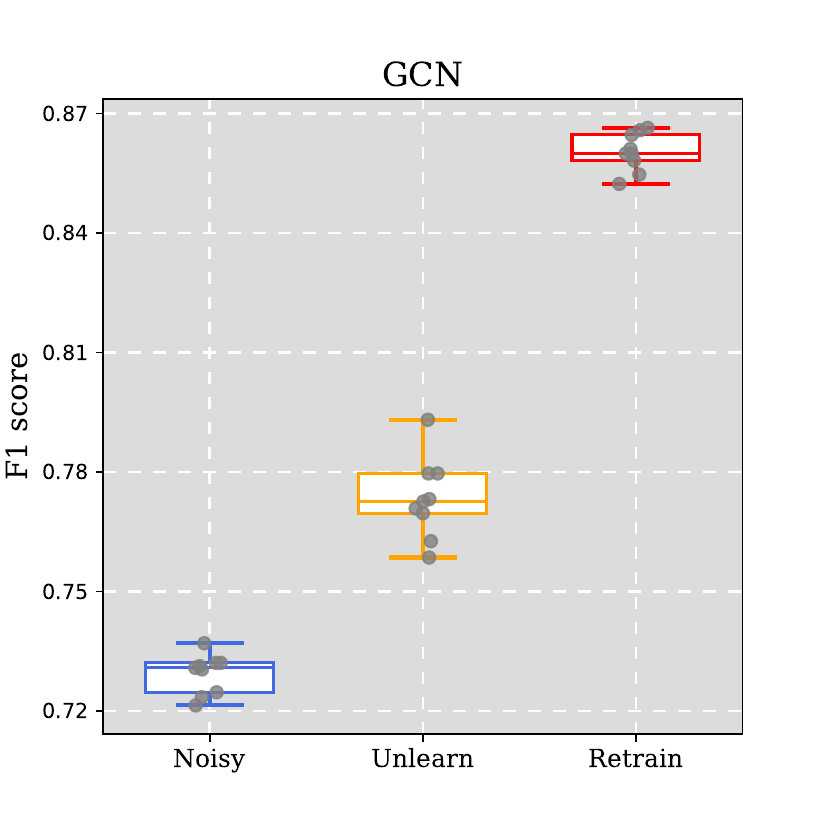}
        \put(-120,30){\rotatebox{90}{\textbf{(b) Citeseer}}}
    \end{minipage}
    \hfill
    \begin{minipage}[b]{0.245\textwidth}
        \centering
        \includegraphics[height = 0.17\textheight, width=1.14\textwidth]{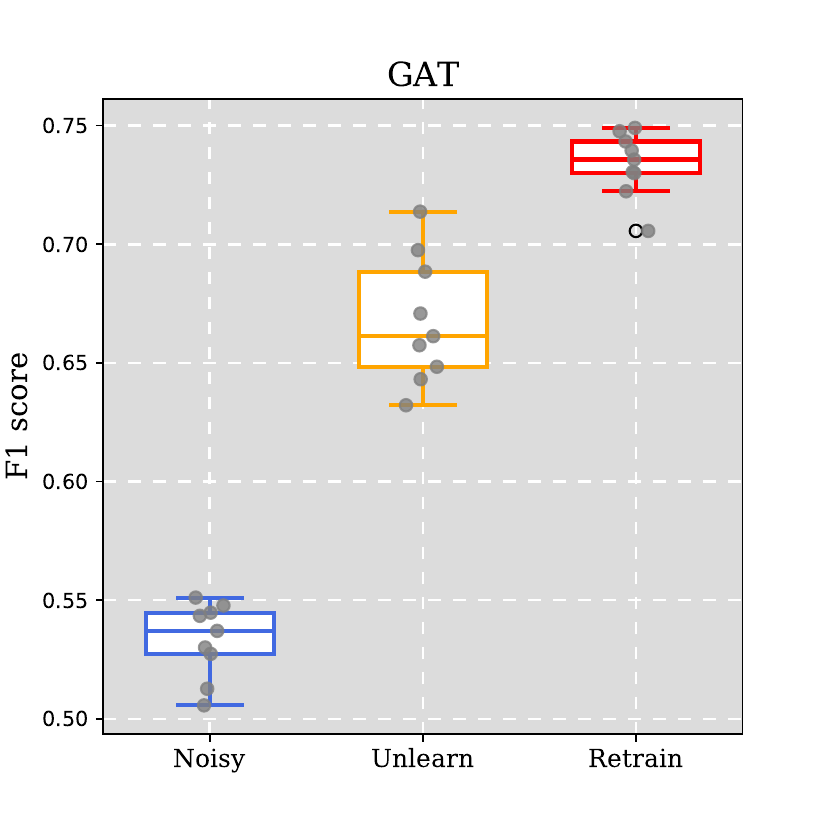}
    \end{minipage}
    \hfill
    \begin{minipage}[b]{0.245\textwidth}
        \centering
        \includegraphics[height = 0.17\textheight, width=1.14\textwidth]{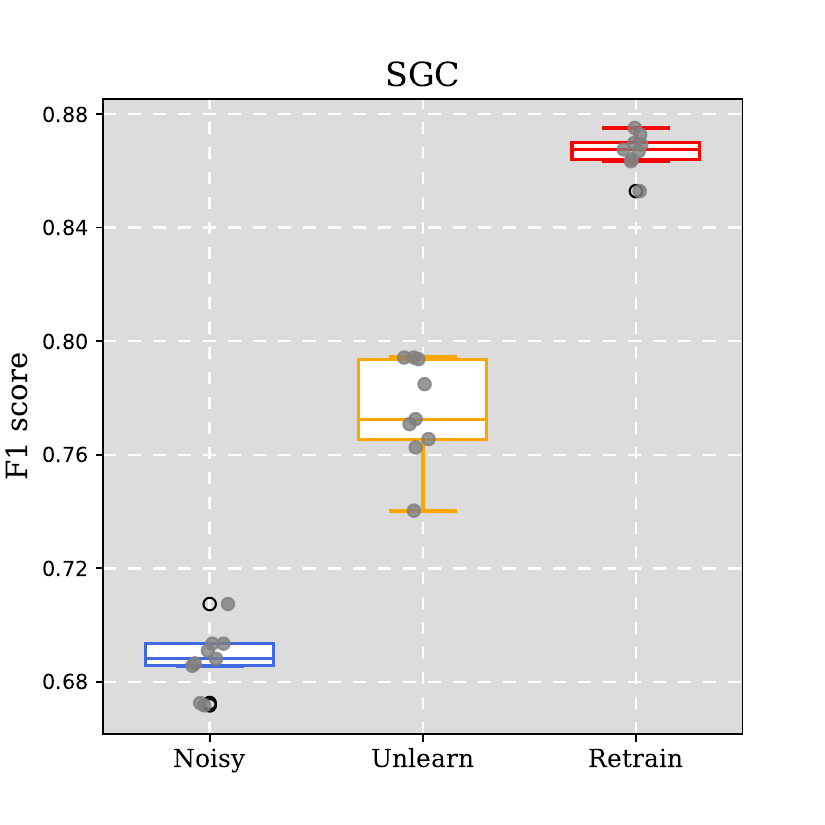}
    \end{minipage}
    \hfill
    \begin{minipage}[b]{0.245\textwidth}
        \centering
        \includegraphics[height = 0.17\textheight, width=1.14\textwidth]{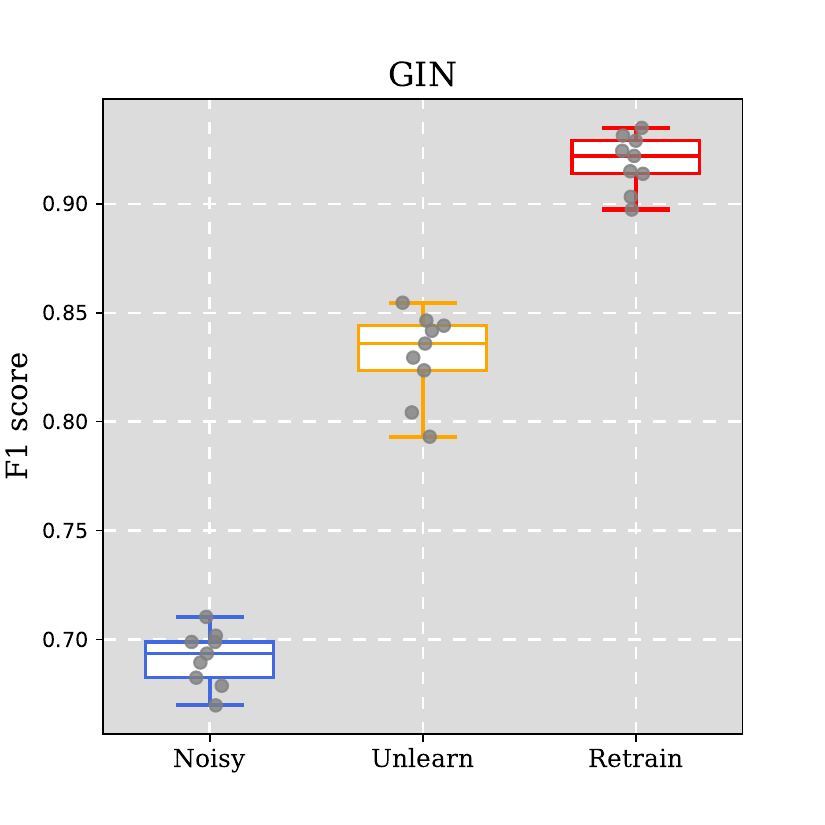}
    \end{minipage}
    \caption{Unlearning efficacy box diagram. Grey data points are the statistics of 10 repeated results.}
    \label{fig:efficacy}
\end{figure}

We find that our proposed method can effectively improve the performance of the original model output representation on downstream tasks on different graph models as shown in Fig.~\ref{fig:efficacy}.

\vspace{-10pt}

\subsection{Ablation Study and Varying Unlearning Ratio} \label{sec:anchor}
We conduct ablation studies to illustrate the Range-Null Space Decomposition's (RND) impact. 

\begin{wrapfigure}{r}{0.63\textwidth}
    \vspace{-10pt}
    \centering
    \includegraphics[height = 0.215\textheight, width = 0.63\textwidth]{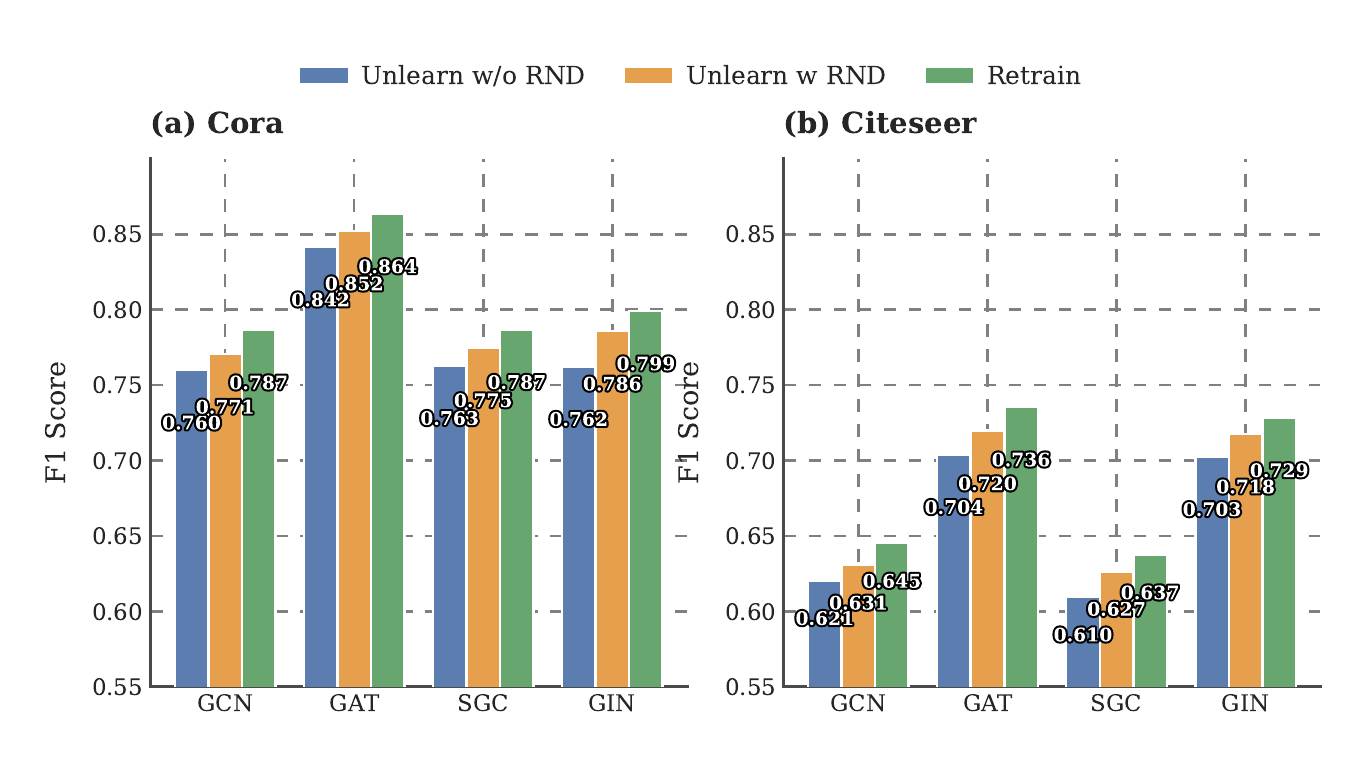}
    \caption{Ablation study of RND on Cora and Citeseer.}
    \vspace{-10pt}
\end{wrapfigure}

\textbf{Ablation Study.} The experiments on Cora and Citeseer show that our method with RND achieves closer utility with retraining, that is, the F1-score of multi-label node classification. In order to show the effectiveness of the RND clearly, the unlearn ratio is set to be 30\%. At the same time, its performance is consistently better than that without RND. This shows that the linear inverse constraints lead to less loss of useful information in the unlearning process.

\begin{figure}[htbp]
    \centering
    \vspace{-10pt}
    \begin{minipage}[b]{0.245\textwidth}
        \centering
        \includegraphics[height = 0.16\textheight, width=1.14\textwidth]{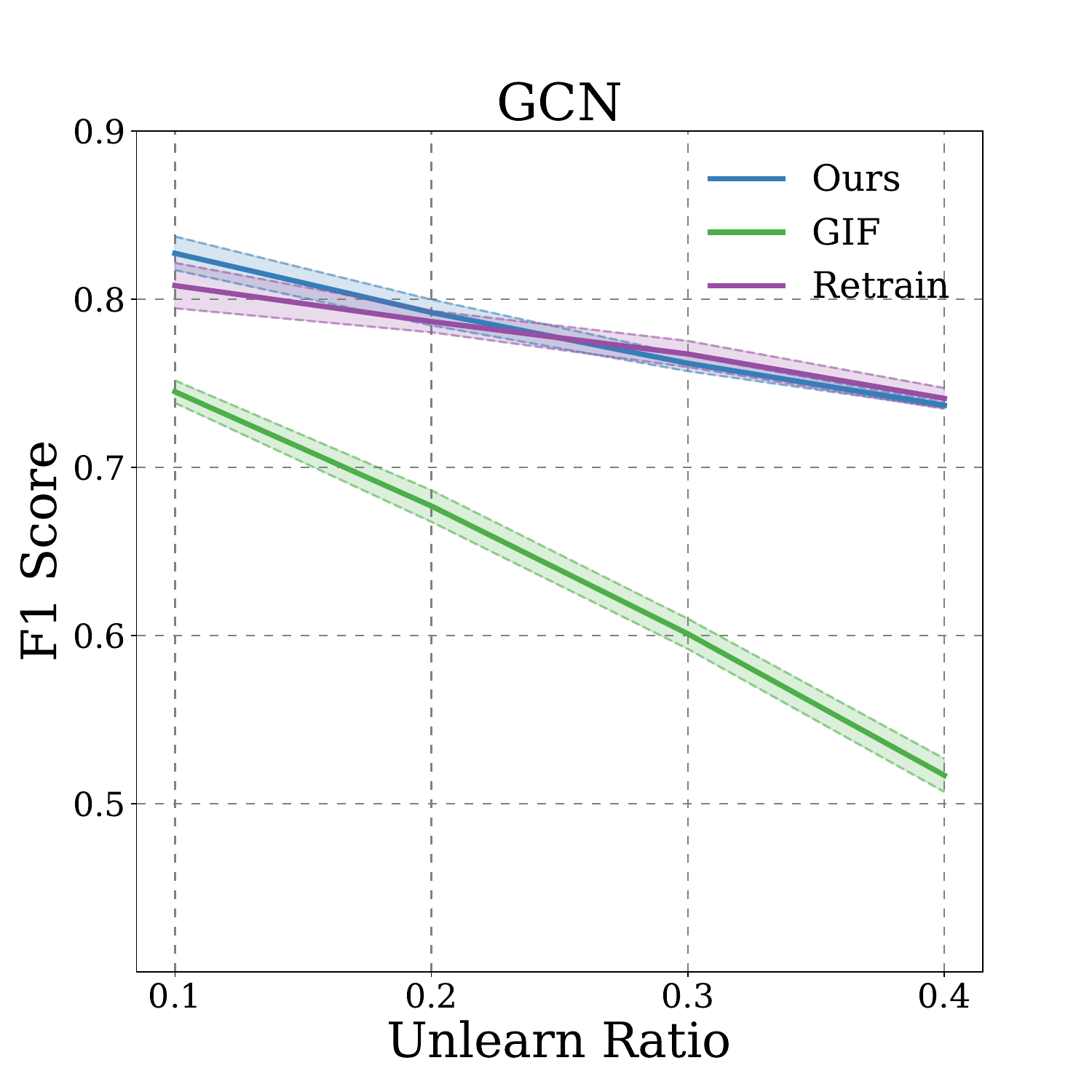}
        \put(-120,30){\rotatebox{90}{\textbf{(a) Cora}}}
    \end{minipage}
    \hfill
    \begin{minipage}[b]{0.245\textwidth}
        \centering
        \includegraphics[height = 0.16\textheight, width=1.14\textwidth]{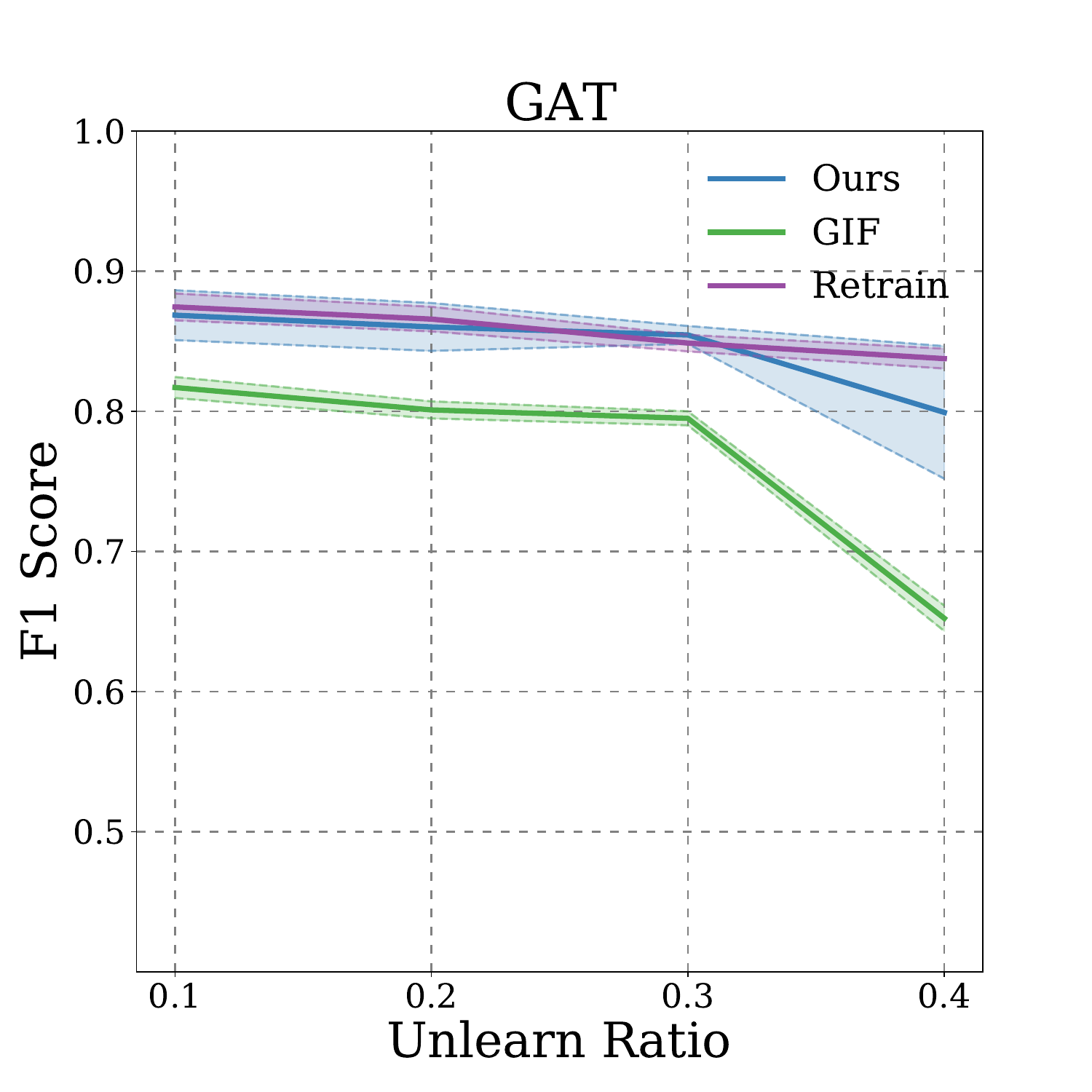}
    \end{minipage}
    \hfill
    \begin{minipage}[b]{0.245\textwidth}
        \centering
        \includegraphics[height = 0.16\textheight, width=1.14\textwidth]{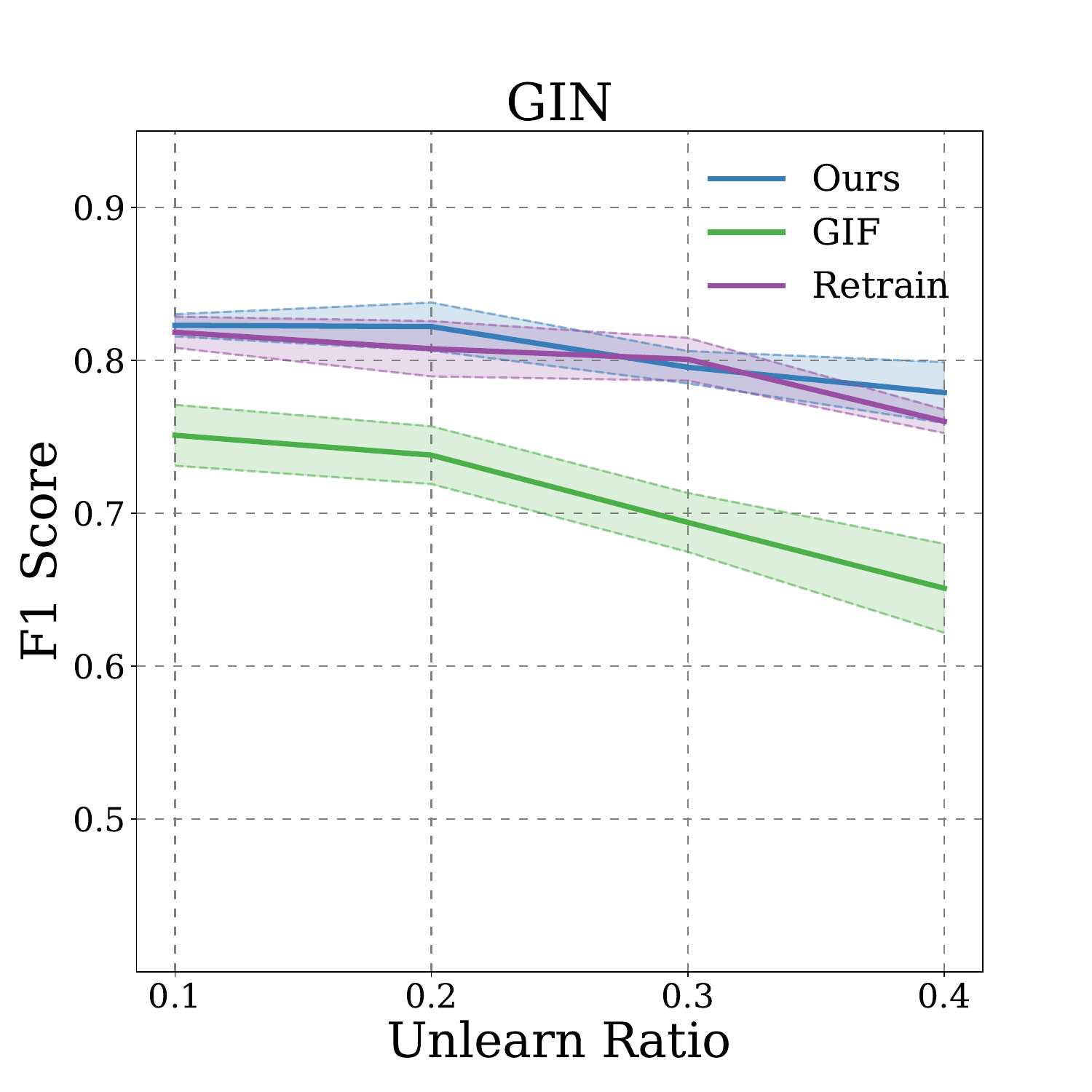}
    \end{minipage}
    \hfill
    \begin{minipage}[b]{0.245\textwidth}
        \centering
        \includegraphics[height = 0.16\textheight, width=1.14\textwidth]{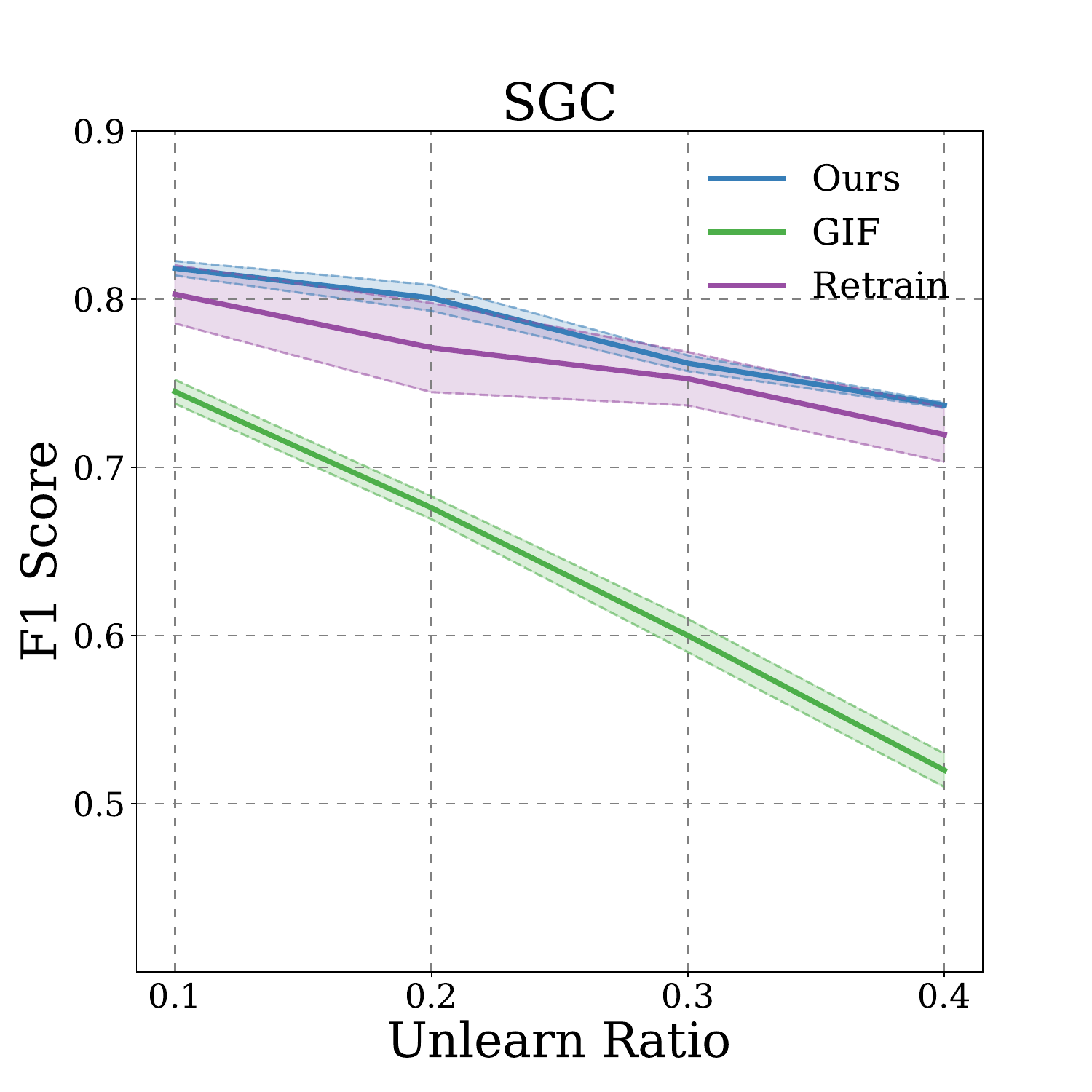}
    \end{minipage}
    \vspace{-10pt}
    \begin{minipage}[b]{0.245\textwidth}
        \centering
        \includegraphics[height = 0.16\textheight, width=1.14\textwidth]{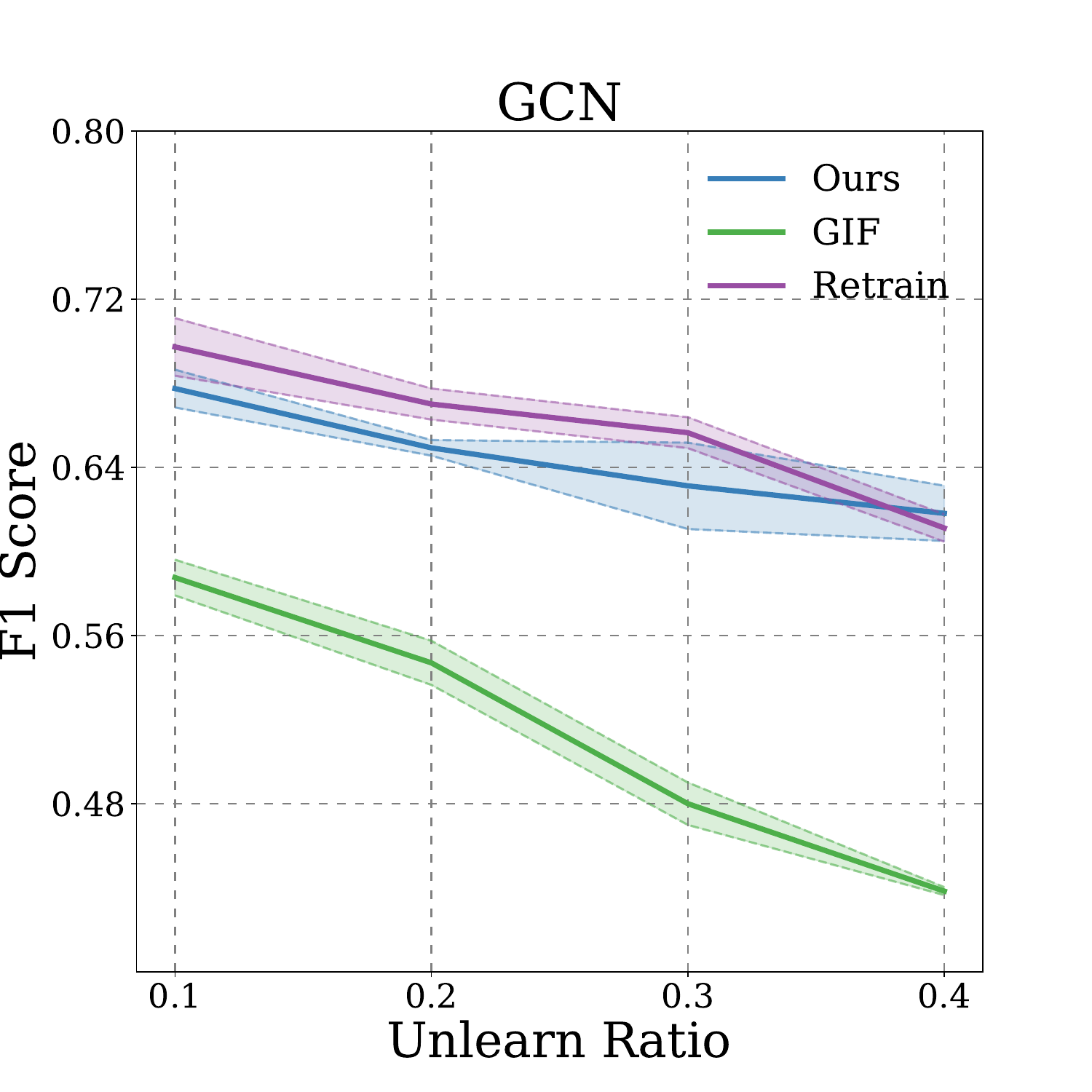}
        \put(-120,30){\rotatebox{90}{\textbf{(b) Citeseer}}}
    \end{minipage}
    \hfill
    \begin{minipage}[b]{0.245\textwidth}
        \centering
        \includegraphics[height = 0.16\textheight, width=1.14\textwidth]{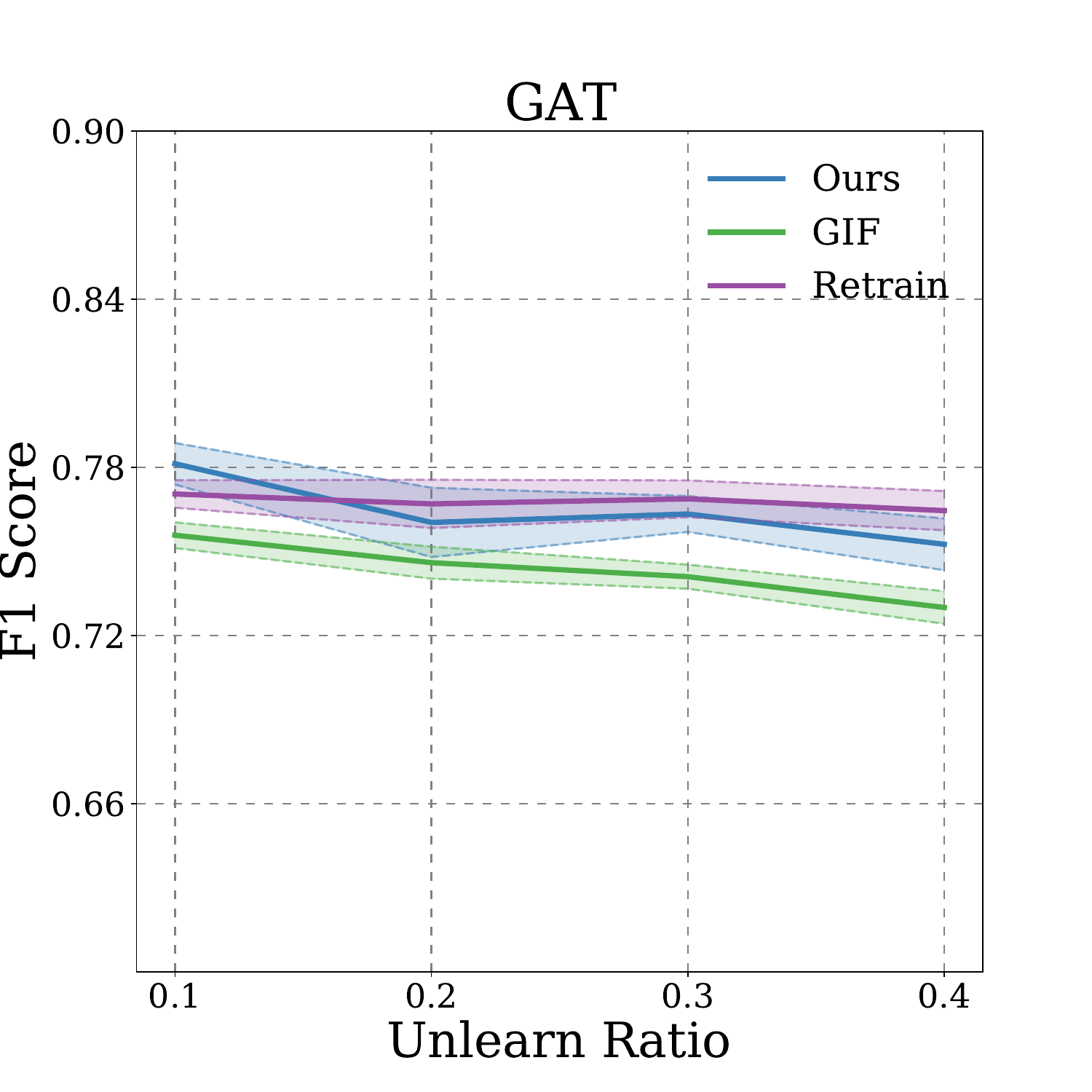}
    \end{minipage}
    \hfill
    \begin{minipage}[b]{0.245\textwidth}
        \centering
        \includegraphics[height = 0.16\textheight, width=1.14\textwidth]{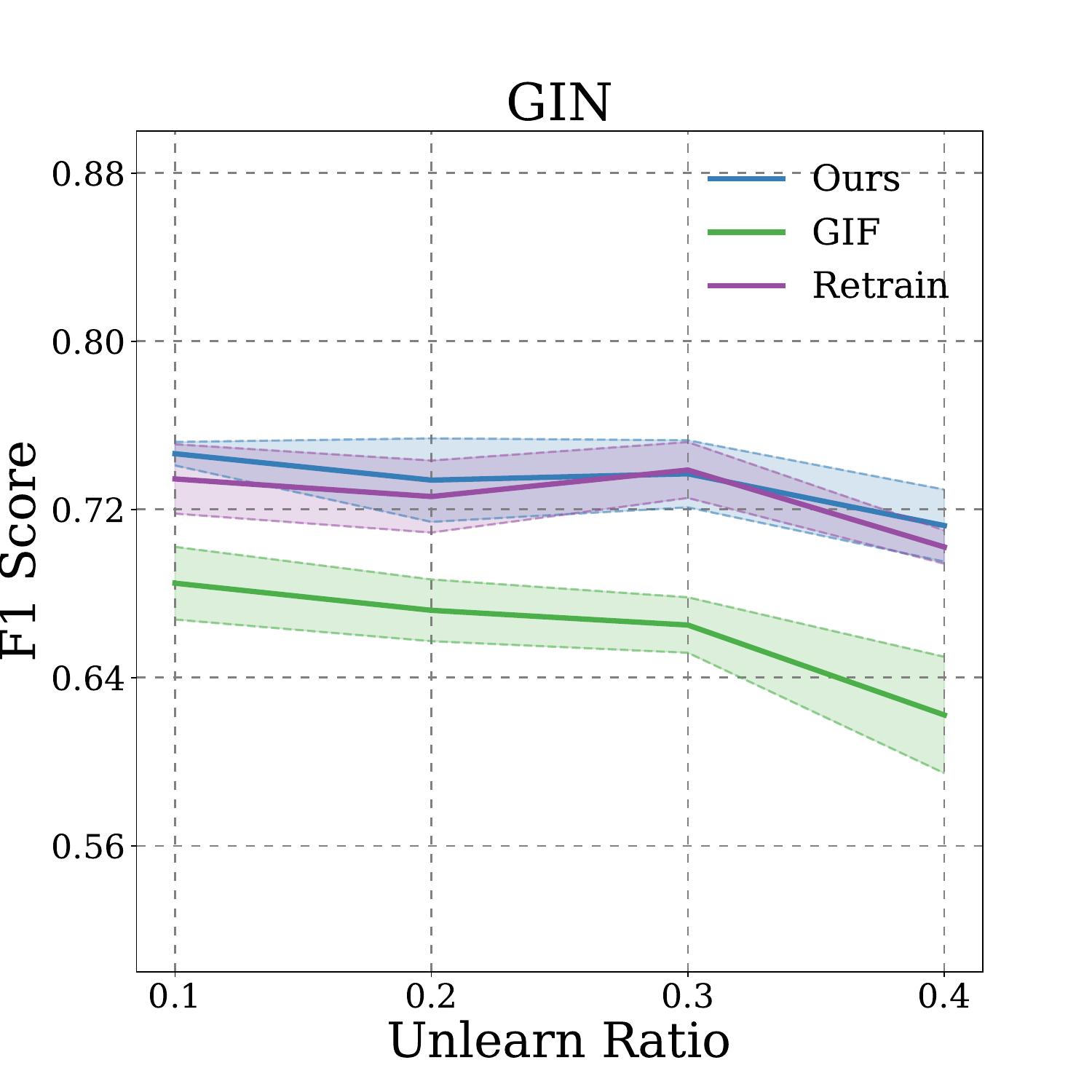}
    \end{minipage}
    \hfill
    \begin{minipage}[b]{0.245\textwidth}
        \centering
        \includegraphics[height = 0.16\textheight, width=1.14\textwidth]{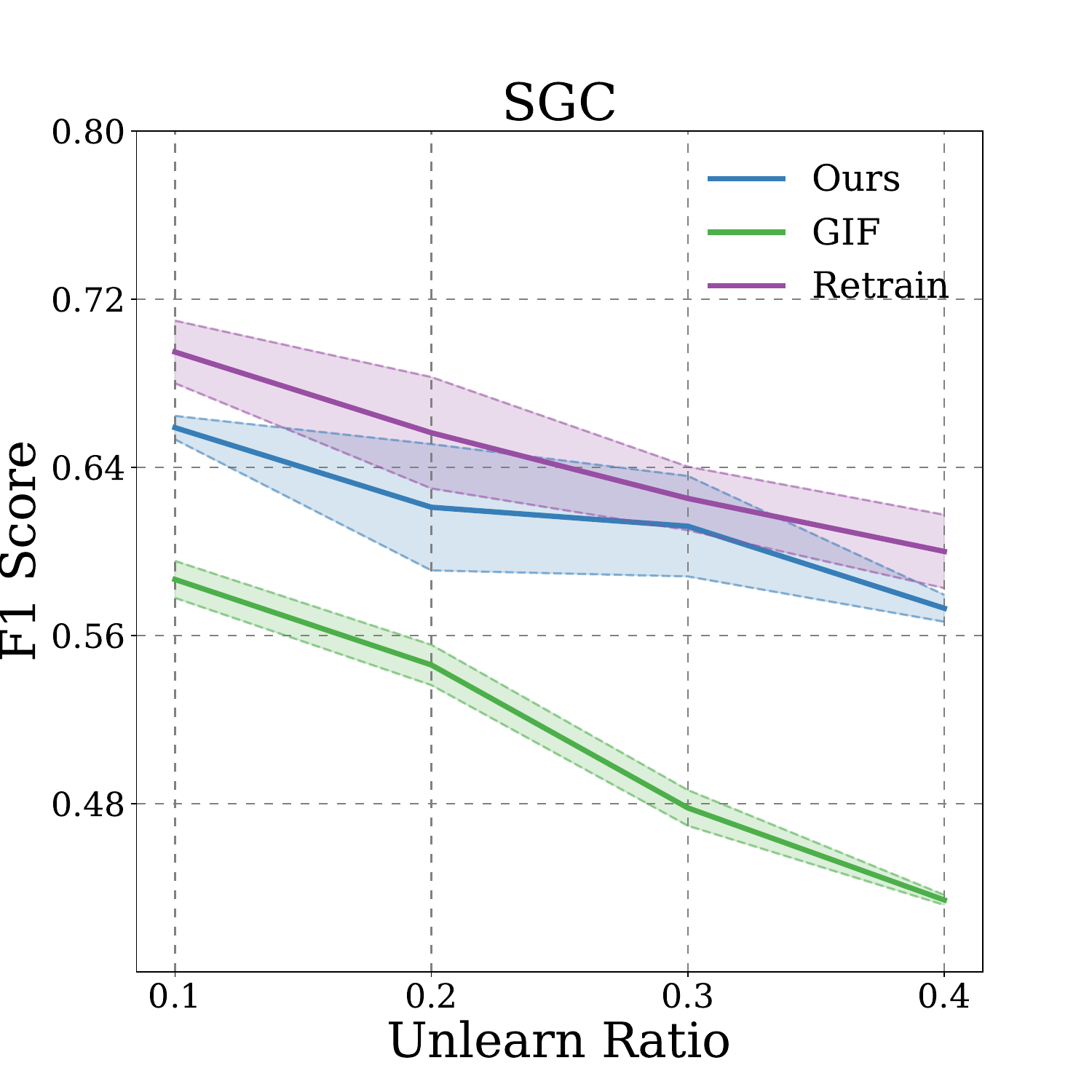}
    \end{minipage}
    \caption{Varying unlearn ratios comparison with GIF and Retrain. Dotted line indicates variances.} 
    \label{fig:varying_ratio}
\end{figure}

\textbf{Varying Unlearning Ratio.} We conduct experiments on Cora and Citeseer under unlearning ratios of 10\%, 20\%, 30\%, and 40\%. It is noted that the model utility of Retrain would decrease slowly and evenly with the decrease of the unlearned proportion. The blue broken-line in Fig.~\ref{fig:varying_ratio} shows our method's performance. It is consistent with the results of Retrain (the perple broken-line), both in variance and mean. Correspondingly, the GIF, as shown by the green broken line, has an obvious gap with Retrain. Its performance drops rapidly with the increase of unlearning ratio. While our method is more robust on different proportions of forgetting requirements.

\subsection{Performance of Privacy Protection Through Membership Inference Attack}
Previous works suggested that unlearning would introduce additional privacy budget \cite{chen2021machine}. Attackers obtain the differences in output embeddings of the GNNs before and after unlearning, respectively, find the nodes with higher differences in representation, and speculate them as unlearned nodes. 

\begin{table*}[htbp]
\centering
\caption{Attack AUC of membership inference against our method ($\mathcal{A}_{\text{I}}$) and retraining ($\mathcal{A}_{\text{II}}$).}
\setlength{\tabcolsep}{16pt}  
\renewcommand\arraystretch{1}\small
\begin{tabular}{c|cc|cc|cc}
\hline
\multirow{2}{*}{Models} & \multicolumn{2}{c}{Cora}               & \multicolumn{2}{|c|}{Citeseer}           & \multicolumn{2}{c}{CS}   \\
                          & $\mathcal{A}_{\text{I}}$   & $\mathcal{A}_{\text{II}}$   & $\mathcal{A}_{\text{I}}$   & $\mathcal{A}_{\text{II}}$  & $\mathcal{A}_{\text{I}}$   & $\mathcal{A}_{\text{II}}$   \\ \hline
\multirow{1}{*}{GCN}      & 0.5014          & 0.5088          & 0.5010    &  0.5202  & 0.4988          & 0.4999         \\ \hline
\multirow{1}{*}{GAT}      & 0.4857          & 0.5209          &       0.4916    & 0.5738         & 0.5015          & 0.5081         \\ \hline
\multirow{1}{*}{SGC}      & 0.4825          & 0.5045          &       0.4963    & 0.5110         & 0.4994          & 0.5026         \\ \hline
\multirow{1}{*}{GIN}      & 0.5129          & 0.5028          &       0.4987    &0.5315          & 0.5002          & 0.5050         \\ \hline
\end{tabular}
\label{tab:exp_attack}
\end{table*}

Note that our method itself is an embedding modification (i.e., noise superposition) based on the originals, and experiments show that all nodes are assigned to relatively uniform noise, which is also the difference that the attacker is trying to obtain. For this reason, as shown in the Table.~\ref{tab:exp_attack}, our method performs well under membership inference attacks. Under various backbones and datasets, the AUC values are close to 0.5 consistently. It shows that the probability of attack success is close to a random guess. The unlearned or retained node embeddings differences are indistinguishable, thus our method's privacy overhead is limited. On the other hand, experiments show that privacy sometimes leaks while retraining, as GAT's AUC on Citeseer deviates from 0.5 by 0.07.

\subsection{Unlearning Visualization} \label{sec:visual}

\begin{wrapfigure}{r}{0.5\textwidth}
    \centering
    \vspace{-10pt}
    \begin{minipage}[b]{0.245\textwidth}
        \centering
        \includegraphics[height = 0.14\textheight, width=1.0\textwidth]{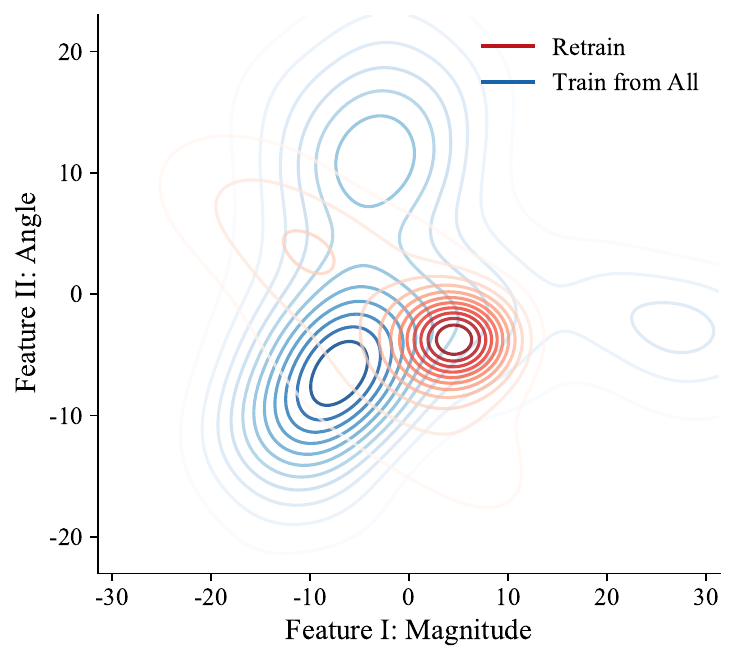}
        \put(-80,80){{\textbf{(a)}}}
    \end{minipage}
    \hfill
    \begin{minipage}[b]{0.245\textwidth}
        \centering
        \includegraphics[height = 0.14\textheight, width=1.0\textwidth]{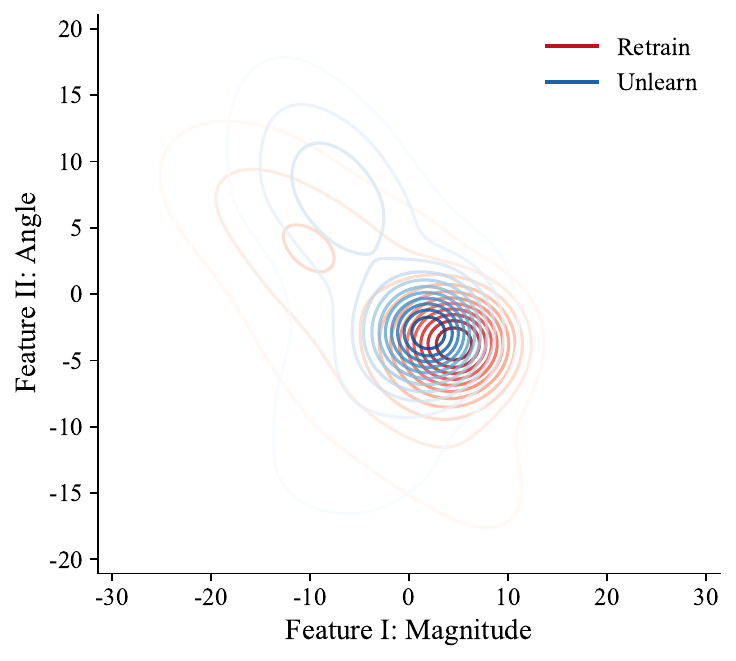}
        \put(-80,80){{\textbf{(b)}}}
    \end{minipage}
    \vspace{-10pt}
    \caption{Embedding distribution under SGC unlearning 30\% nodes.}
    \label{fig:visual}
\end{wrapfigure}
We employ 2D kernel density estimation to calculate the probability density function (PDF) $f_V$ for embedding set $V = {(Mag_i, Ang_i)}_{i=1}^n$. We use Gaussian kernel for PDF estimation, expressed as Eq.~\ref{PDF}. In each domain(training from all, retrain, and unlearn), we independently derive the PDFs with bandwidth $h=1$ as illustrated in Fig.~\ref{fig:visual}. We verify that the embedding distribution of retrain overlaps in space but differs in density with training from all as Fig.~\ref{fig:visual}(a). Our method, as shown in Fig.~\ref{fig:visual}(b), can produce the embedding distributions consistent with that of retraining. 
\begin{equation}
	\label{PDF}
	f_V(Mag, Ang)=\frac{1}{nh^2}\sum_{i=1}^{n} \frac{1}{2\pi} \exp \left\{-\frac{1}{2h^2}[(Mag-Mag_i)^2-(Ang-Ang_i)^2] \right\}
\end{equation}

\section{Conclusion}
We found that previous graph unlearning methods suffer from a great performance decline in node unlearning tasks. Our goal is to address that problem without assuming gradient convergence and achieve better model utility. A novel graph node unlearning method is proposed. The core idea is to reverse the process of aggregation in GNN training by embedding reconstruction and to adopt a Range-Null Space Decomposition for the interaction correction. Experimental results on multiple representative datasets demonstrate the effectiveness of our proposed approach. This work also has important implications for real-world applications, such as adversarial data removal in the training process, etc.

\newpage
\begin{ack}
Thanks to the co-authors for their review and dedication to the paper.

\end{ack}

\bibliographystyle{plain}
\bibliography{reference.bib}
\clearpage
\newpage

\appendix
\onecolumn
\newpage

\begin{appendix}
	\thispagestyle{plain}
	\begin{center}
		{\Large \bf Appendix}
	\end{center}
\end{appendix}

\section{Detailed Related Work}
\subsection{Machine Unlearning Appendix} 
The latter is designed for more efficient unlearning without retraining through fine-tuning the existing model parameters. Adapting the influence function \cite{koh2017understanding} in the unlearning tasks, Guo et al. \cite{guo2020certified} proposed to unlearn by removing the influence of the deleted data on the model parameters. Specifically, they used the deleted data to update models by performing a Newton step to approximate the influence of the deleted data and remove it, then they introduced random noise to the training objective function to ensure the certifiability. Unrolling SGD \cite{thudi2022unrolling} proposes a regularizer to reduce the ‘verification error’, which is an approximation to the distance between the unlearned model and a retrained-from-scratch model. The goal is to make unlearning easier in the future. Langevin Unlearning \cite{chienlangevin} leverages the Langevin dynamic analysis for the machine unlearning problem.

\subsection{Graph Unlearning Appendix}

(1) The Shards-based method: GraphEraser \cite{chen2022graph} and GUIDE \cite{wang2023inductive} extend the shards-based idea to graph-structured data, which offers partition methods to preserve the structural information and also designs a weighted aggregation for inference. Moreover, GraphRevoker \cite{zhang2024graph} utilized a property-aware sharding method and contrastive sub-model aggregation for efficient partial retraining and inference. (2) The IF-based method: like CGU \cite{chien2022certified}, GIF \cite{wu2023gif}, CEU \cite{wu2023certified}, IDEA \cite{dong2024idea}, and GST \cite{pan2023unlearning} extends the influence-function method and proposes a similar formula for edge and node unlearning tasks on the graph model and further analyzes the theoretical error bound of the estimated influences under the Lipschitz continuous condition and loss convergence condition. (3) Learning-based method: GNNDelete \cite{chenggnndelete} bounding edge prediction through a deletion operator and pays little attention to node embeddings equivalence. MEGU \cite{li2024towards} achieved effective and general graph unlearning through a mutual evolution design. (4) Others: Projector \cite{cong2023efficiently} provides closed-form solutions with theoretical guarantees. 


\begin{table*}[htbp]
\centering
\caption{F1-score $\pm$ STD comparison under the standard setting of transductive node classification task with node unlearning request. The highest results are highlighted in \sethlcolor{gray!50}\hl{bold} while the second-highest results are marked with \sethlcolor{lightgray!80}\hl{grey}. Closer to \sethlcolor{yellow!30}\hl{Retrain}'s F1-score means better performance, and OOT indicates 'Out Of Time'.}
\setlength{\tabcolsep}{1.8pt}  
\renewcommand\arraystretch{1}\small
\begin{tabular}{cc|c|c|c|c|c|c}
\hline
 & \multirow{2}{*}{Strategy} & \multicolumn{2}{c}{Cora}               & \multicolumn{2}{c}{Citeseer}           & \multicolumn{2}{c}{CS}   \\ \cline{3-8}
                          &                           & F1-score                   & RT (second)        & F1-score   & RT (second)  & F1-score  & RT (second)        \\ 
                          \hline
                          & Retrain                   & \cellcolor{yellow!30}0.8357$\pm$0.0161          & \cellcolor{yellow!50}5.96          & \cellcolor{yellow!30}0.6756$\pm$0.0132          & \cellcolor{yellow!50}7.69          & \cellcolor{yellow!30}0.8942$\pm$0.0147          & \cellcolor{yellow!50}102.46         \\ \hline
                          & GraphEraser               & 0.8114$\pm$0.0100          & 105.51          & 0.7357$\pm$0.0125          & 135.65          & 0.9124$\pm$0.0008          & 1323.26          \\
                          & GUIDE                     & 0.7389$\pm$0.0218          & 44.80          & 0.6350$\pm$0.0071          & 57.60          & 0.8696$\pm$0.0015          & 2128.36         \\
                          & GraphRevoker              & 0.8109$\pm$0.0109          & 30.54          & 0.7345$\pm$0.0061          & 39.27          & 0.9126$\pm$0.0010          & 793.96         \\ \hline
                          & GIF                       & 0.8175$\pm$0.0109          & 0.15          & 0.6258$\pm$0.0067          & \cellcolor{lightgray!80}0.18          & 0.9187$\pm$0.0022          & \cellcolor{lightgray!80}11.12          \\
                          & CGU                       & 0.8637$\pm$0.0078          & 82.85          & 0.7562$\pm$0.0041          & 106.53        &  OOT          & OOT          \\
                          & ScaleGUN                  & 0.7882$\pm$0.0014          & 1.09          & 0.7342$\pm$0.0013          & 1.31          & 0.9144$\pm$0.0009          & 55.60         \\  
                          & IDEA                      & 0.8771$\pm$0.0025          & \cellcolor{lightgray!80}0.14          & 0.6366$\pm$0.0049          & \cellcolor{gray!50}\textbf{0.17}          & \cellcolor{gray!50}\textbf{0.8947$\pm$0.0022}          & 33.35          \\ \hline
                          & GNNDelete                 & 0.7478$\pm$0.0549          & 0.95          & \cellcolor{lightgray!80}0.6426$\pm$0.0382          & 1.14          & 0.7626$\pm$0.0273          & 122.32          \\
                          & MEGU                      & \cellcolor{gray!50}\textbf{0.8268$\pm$0.0156}          & 0.95          & 0.6360$\pm$0.0111          & 1.15          & 0.9169$\pm$0.0008          & 123.35          \\ \hline
                          & Projector                  & 0.8679$\pm$0.0237          & 5.86          & 0.7700$\pm$0.0065          & 7.03          & 0.8840$\pm$0.0063          & 88.97          \\ \hline
                          & Ours                       & \cellcolor{lightgray!80}0.8518$\pm$0.0143 & \cellcolor{gray!50}\textbf{0.14} & \cellcolor{gray!50}\textbf{0.6645$\pm$0.0068} & 0.40 & \cellcolor{lightgray!80}0.9013$\pm$0.0089 & \cellcolor{gray!50}\textbf{0.96} \\ \hline
\end{tabular}
\label{tab:exp_SGC_all}
\end{table*}

\section{Comprehensive Comparison with Existing Method}\label{appendix:comp}
Previous studies on GU have often employed varying dataset splits, different GNN backbones, and inconsistent unlearning request configurations, hindering direct comparisons between different methods. According to the existing benchmark OpenGU \cite{fan2025opengu}, we use datasets split into 80\% for training and 20\% for testing. For unlearning requests, 10\% of nodes are selected for removal. Regarding backbone selection, we leverage SGC as a representative of decoupled GNNs for the node unlearning task. We report the mean performance and standard deviation over 10 runs, ensuring consistency and reliability in the evaluation.

We compare our method with four types existing methods as Tab.~\ref{tab:exp_SGC_all}. \textit{The Shards-based method}: the additional subgraph partition cost is introduced, and the unlearned node subgraph retraining process also constitutes an important part of its time cost. Compared with training from scratch, their efficiency is very low and lacks model utility. This kind of method often sacrifices a part of graph structure in the process of community division, which leads to the forgetting of some useful information and the degradation of its performance.

\textit{The IF-based method}: This types of method achieves comparable performance with retraining, this kind of method is the most practical one according to experimental verification. IDEA achieves the best performance in CS dataset, while lack of efficiency compared with ours. On the small-scale graph, its efficiency is competitive. However, due to the iterative approximation of hessian matrix of parameters, this kind of method will increase in time complexity at $O(n^2)$ with the expansion of graph nodes number. Although GIF has a gap with IDEA in performance, it has good scalability for it could be extended under multiple backbones and multiple task settings \cite{fan2025opengu}.

\textit{The Learning-based method}: GNNdelete is more applicable in linear GNN models while MEGU achieves the best performance in Cora dataset. The latter introduced the concept of High-influence nodes for optimization, thus complicated interactions in large scale graph hinders the selection and optimization of such nodes. MEGU method is not good in model efficiency, compared with IF-based method. 

\textit{Others}: Projector is more specialized, limiting its scalability and generalization \cite{fan2025opengu}. The comparison of experimental data also shows that its performance and efficiency are not dominant compared with other types methods.

Our method has achieved good performance in multiple datasets and experimental settings. Compared with retraining, the utility of the our model has not been greatly affected. It has a very obvious advantage in efficiency, especially on large-scale graphs CS. This fully demonstrates the practicability of this method. 

\section{Method Extension}
\subsection{Low Ratio Removal}
As we all know, machine learning methods rely on a sufficient training data to take effect. As we model unlearning as a data-driven machine learning problem, to provide sufficient training data cannot be ignored. Specifically, sufficient unlearned nodes for embeddings reconstruction towards training of the embeddings modification model is important. When the number of nodes to be forgotten is very small, node interactions are hard to learn according to Eq.~\ref{L_reconst}. Considering that our unlearning method training and inference are strongly decoupled, all nodes besides unlearned nodes also satisfy the interaction between nodes. The loss function of interaction learning of $\mathcal{L}^{\text{Inter}}$ could be extended as follows $\mathcal{L}^{\text{Inter+}}$:
\begin{equation}
\label{L_reconst_new}
\mathcal{L}^{\text{Inter+}}=\frac{1}{|m|}\sum_{m \in \mathcal{U\cup R}} KL\left(\text{Norm}[\bm{h}_{e_m}^{k-1}],\text{Norm}[\sum_i{f_2(i,m)}]\right)
\end{equation}
This $\mathcal{L}^{\text{Inter+}}$ loss function is attached to our experiment for implementation, and the results in the main body are obtained.

\subsection{High Ratio Removal}

When the ratio of forgotten nodes is very large, like higher than 30\%, the graph structural factors play a major role in embedding distribution, as our experimental experience. 

In order to adapt to this scene, we modified the formula of $\tilde{\bm{h}}^{k}_{e_i}$ to achieve better forgetting. With the help of the training graph $G'=G/\Delta G$ after deleting the unlearned nodes, we get the middle embedding through direct inference $\overline{\bm{h}}^{k}_{e_i}=\text{GNN}_0(G')$, where $\text{GNN}_0(\cdot)$ is the model before unlearning. Thus, $\tilde{\bm{h}}^{k}_{e_i}$ is given by:
\begin{equation}
\tilde{\bm{h}}^{k}_{e_i}=\overline{\bm{h}}^{k}_{e_i}-\gamma\sum_{j \in \mathcal{U}}f_1(j,i)
\label{eq:new_unlearn}
\end{equation}

By the way, the \textbf{Local search loss} is also transformed into:
\begin{equation}
\mathcal{L}^{\text{Local+}}=\sum_{p} KL(\text{Norm}[\overline{\bm{h}}^{k}_{e_i}],\text{Norm}[\tilde{\bm{h}}^{k}_{e_p}]).
\end{equation}

\begin{wrapfigure}{r}{0.5\textwidth}
    \centering
    \vspace{-10pt}
    \begin{minipage}[b]{0.245\textwidth}
        \centering
        \includegraphics[height = 0.145\textheight, width=1.0\textwidth]{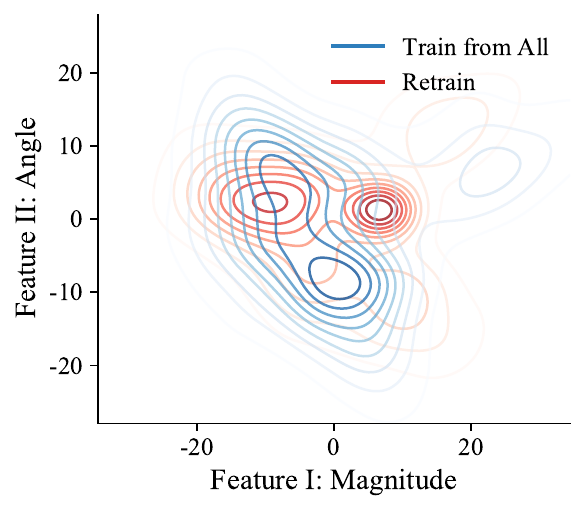}
        \put(-80,90){{\textbf{(a)}}}
    \end{minipage}
    \hfill
    \begin{minipage}[b]{0.245\textwidth}
        \centering
        \includegraphics[height = 0.145\textheight, width=1.0\textwidth]{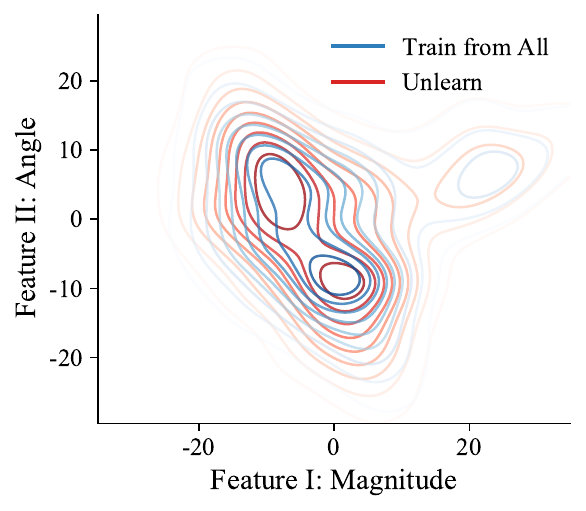}
        \put(-80,90){{\textbf{(b)}}}
    \end{minipage}
    \vspace{-10pt}
    \caption{Embedding distribution under SGC unlearning 40\% nodes.}
    \label{fig:visual_new}
\end{wrapfigure}

At this time, if we still employ embedding ${\bm{h}}^{k}_{e_i}=\text{GNN}_0(G)$ and obtain $\tilde{\bm{h}}^{k}_{e_i}$ by Eq.~\ref{equation8}, the produced embedding distribution is not equivalent to retraining. As shown in Fig.~\ref{fig:visual_new}(a), the embedding distribution of retrain is far different from training from all. The PDFs of the two have clearly deviated in angle. While the obtained unlearned distribution almost overlaps with that of training from all in Fig.~\ref{fig:visual_new}(b). 

It is difficult to complete the characterization distribution with a huge difference before and after retraining through a small disturbance. Therefore, it is better to use a middle result considering structural factors in this scene.

\subsection{Edge Unlearning}

we simply use an MLP layer named $MLP_1$ to represent the interaction to be removed in $\bm{h}^{k}_{e_i}$ in the $k$-th layer between remained node $e_i$ and unlearned node $e_j$. It should be noted that when the above $f_1$ is subtracted from $\bm{h}^{k}_{e_i}$, the influence of an edge $e_j\to e_i$ is eliminated actually. 

Thus, given unlearning edge target as $\mathcal{U}_e$, edge unlearning may be extended in inference as:

\begin{equation}
\tilde{\bm{h}}^{k}_{e_i}=\bm{h}^{k}_{e_i}-\gamma\sum_{<j,i> \in \mathcal{U}_e}f_1(j,i)
\end{equation}

\end{document}